\documentclass[conference]{ieeeconf}
\IEEEoverridecommandlockouts
\usepackage{times}
\usepackage{graphicx}
\usepackage{multicol}
\usepackage[bookmarks=true]{hyperref}
\usepackage{array}
\usepackage[detect-all]{siunitx}
\usepackage{amsmath}
\DeclareMathOperator*{\argmax}{arg\,max}
\DeclareMathOperator*{\argmin}{arg\,min}

\usepackage{color}

\newcommand{\lpap}{$\textrm{LP}_0\textrm{AP}_1$}

\begin{document}

\title{Learning to Fold Real Garments with One Arm: \\ A Case Study in Cloud-Based Robotics Research} % Benchmarking with Cloud-Based Robotics Research?

% You will get a Paper-ID when submitting a pdf file to the conference system
\author{Ryan Hoque$^{*,1}$, Kaushik Shivakumar$^{*,1}$, Shrey Aeron$^1$, Gabriel Deza$^1$, Aditya Ganapathi$^1$, \\ Adrian Wong$^2$, Johnny Lee$^2$, Andy Zeng$^2$, Vincent Vanhoucke$^2$, Ken Goldberg$^1$%
\thanks{$^*$ Equal contribution}
\thanks{$^{1}$AUTOLAB at University of California, Berkeley}
\thanks{$^{2}$Robotics at Google}
}

%\author{\authorblockN{Michael Shell}
%\authorblockA{School of Electrical and\\Computer Engineering\\
%Georgia Institute of Technology\\
%Atlanta, Georgia 30332--0250\\
%Email: mshell@ece.gatech.edu}
%\and
%\authorblockA{Twentieth Century Fox\\
%\and
%\authorblockN{James Kirk\\ and Montgomery Scott}
%\authorblockA{Starfleet Academy\\
%San Francisco, California 96678-2391\\

% avoiding spaces at the end of the author lines is not a problem with
% conference papers because we don't use \thanks or \IEEEmembership

%Homer Simpson\authorrefmark{2},
%Montgomery Scott\authorrefmark{3} and
%Eldon Tyrell\authorrefmark{4}}
%Telephone: (800) 555--1212, Fax: (888) 555--1212}
%\authorblockA{\authorrefmark{4}Tyrell Inc., 123 Replicant Street, Los Angeles, California 90210--4321}}

\maketitle

\begin{abstract}
%Autonomous fabric manipulation is a longstanding challenge in robotics, but evaluating progress is difficult due to the cost and diversity of robot hardware. Using Reach, a new cloud robotics platform that enables low-latency remote execution of control policies on physical robots, we evaluate and compare 8 algorithms on the task of folding a crumpled T-shirt with a single robot arm. We develop 4 novel learning-based algorithms that model expert actions, keypoints, reward functions, and dynamic motions, and we benchmark these against learning-free and inverse dynamics algorithms. The entire lifecycle of data collection, model training, and policy evaluation is performed remotely without physical access to the robot workcell. Results suggest a new algorithm combining imitation learning with analytic methods achieves 84\% of human-level performance on the folding task. See \url{https://sites.google.com/berkeley.edu/cloudfolding} for all data, code, models, and supplemental material.
%\todo{include wrinkle score in this somehow?}

Autonomous fabric manipulation is a longstanding challenge in robotics, but evaluating progress is difficult due to the cost and diversity of robot hardware. Using Reach, a cloud robotics platform that enables low-latency remote execution of control policies on physical robots, we present the first systematic benchmarking of fabric manipulation algorithms on physical hardware. %We evaluate and compare 8 algorithms on the task of folding a crumpled T-shirt with a single robot arm. 
We develop 4 novel learning-based algorithms that model expert actions, keypoints, reward functions, and dynamic motions, and we compare these against 4 learning-free and inverse dynamics algorithms on the task of folding a crumpled T-shirt with a single robot arm. The entire lifecycle of data collection, model training, and policy evaluation is performed remotely without physical access to the robot workcell. Results suggest a new algorithm combining imitation learning with analytic methods achieves 84\% of human-level performance on the folding task. See \url{https://sites.google.com/berkeley.edu/cloudfolding} for all data, code, models, and supplemental material.

\end{abstract}

\IEEEpeerreviewmaketitle

\section{Introduction}
\label{sec:introduction}

Reproducibility is the cornerstone of scientific progress. It allows researchers to verify results, assess the state of the art, and build on prior work. Recent advances in computer vision (CV), for instance, were facilitated by the ImageNet Large Scale Visual Recognition Challenge (ILSVRC)~\cite{imagenet}, a benchmark that has become standard in CV literature. 

In robotics, there is no equivalent benchmark because of robots' interaction with physical environments. They are also expensive and vary greatly in their capabilities and morphologies. Accordingly, each research lab has a unique hardware setup, making it difficult to reliably compare results. Such barriers to entry pose a significant obstacle to individuals or institutions who wish to perform robotics research, but lack the resources to do so. Simulation benchmarks \cite{ALFRED, openai_gym, softgym} are one way to address these issues, but the ``reality gap" between simulation and the real world remains prohibitively large \cite{Collins2019QuantifyingTR}.

One alternative is to provide shared access to a remote testbed with standardized hardware via the Cloud. %Several groups have explored this; most recently, teams around the world evaluated their algorithms for dexterous manipulation of rigid objects in the Real Robot Challenge \cite{robotcluster} in 2020 and 2021. 
Although testbeds impose costs to hosts who must set up and maintain hardware, they have the potential to significantly increase reproducibility and accessibility.
%, remote testbeds can simplify hardware usage,  allow robot operation at arbitrary physical distances, and enable the collection of reusable real data. However, the testbeds impose a cost to the maintainer, may require on-site assistance, and rely on low network latency.
In this paper, we describe algorithms and experiments performed entirely remotely using Reach, a prototype hardware testbed from Robotics at Google \cite{reach2022pyreach}. 
% Meanwhile, the Internet has become a mature and ubiquitous technology enabling real-time communication across the globe; the term ``Cloud Robotics" was coined by James Kuffner in 2010 to refer to robots that harness the power of the Internet. As such, new initiatives such as the Real Robot Challenge \cite{robotcluster} and Google Reach \todo{cite} seeks to facilitate reproducibility in robotics by maintaining a cloud robotics testbed. 
Reach consists of several physical robot workcells as well as open source software for remote execution of control policies in real time. Each workcell is configured for a particular benchmark task: one such task is folding a T-shirt with a UR5 robot arm and 3-jaw piSOFTGRIP gripper \cite{piab}.

While folding T-shirts and other garments is a ubiquitous daily task for humans, manipulating fabric remains challenging for robots. Fabric is difficult to model due to its infinite-dimensional state space, complex dynamics, and high degree of self-occlusion. % as well as underactuated dynamics, where the fabric deforms only locally near the robot manipulator. 
Furthermore, accurately simulating gripper contact mechanics and fabric self-collision remains elusive for existing fabric simulators due to challenges in modeling deformation, friction, and electrostatic forces \cite{softgym,seita_fabrics_2020}. Nevertheless, to our knowledge, there has been no systematic benchmarking of algorithms for physical garment folding.

This paper makes the following contributions: (1) four novel learning-based algorithms for the folding task, (2) implementation of and comparison with four additional benchmarks, %(3) a novel evaluation metric for folding, 
and (3) a case study of robotics research performed exclusively using a remotely managed robot workcell. This paper does \textit{not} contribute the design of the Reach cloud robotics platform, which is being developed by a larger team at Google \cite{reach2022pyreach}.

\begin{figure}
    \centering
    \includegraphics[width=0.5\textwidth]{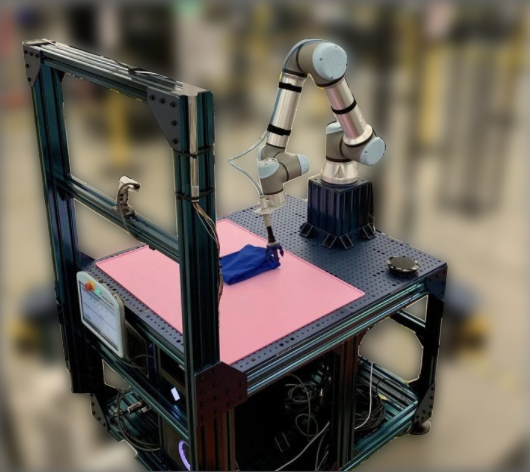}
    \caption{A Reach cloud robotics workcell developed by Robotics and Google.}
    \label{fig:reach-cell}
    \vspace{-0.6cm}
\end{figure}
\section{Related Work}
\label{sec:related-work}

\textbf{Remote Testbeds:} The most similar remote robotics testbed initiative is from Bauer et al. \cite{robotcluster}, who hosted the online ``Real Robot Challenge" in 2020 and in 2021 at Neural Information Processing Systems (NeurIPS). Six robotics groups from around the world were able to access their tri-finger robot \cite{trifinger} remotely via the Internet and evaluate their algorithms on the shared infrastructure. Our study differs from this project in the following ways: (1) they consider dexterous manipulation of rigid objects while we consider deformable object manipulation; (2) they use a custom tri-finger robotic system while we use a UR5 robot arm, standard in industrial settings; and (3) the Real Robot Challenge submissions are either learning-free \cite{benchmarking2022, Chen2021DexterousMP, yoneda2021grasp} or learned only in simulation \cite{mccarthy2021solving, allshire2021transferring}, while we consider learning algorithms trained on real data. Other remote testbeds that do not consider manipulation are the Robotarium \cite{robotarium} for swarm robotics and Duckietown \cite{duckietown} for autonomous driving.

\textbf{Reproducibility in Robotics:} Several other approaches have been proposed for facilitating reproducibility in robotics research. One direction is benchmarking in simulation, where evaluation is inexpensive and reproducible. Simulation environments have been developed for robot locomotion \cite{openai_gym}, household tasks \cite{ALFRED}, and deformable object manipulation \cite{softgym}. While researchers have made significant progress on these benchmarks~\cite{softactorcritic, td3}, especially using reinforcement learning, such advances do not readily transfer to physical robots \cite{Collins2019QuantifyingTR}. Another initiative for improving reproducibility is development of a low-cost open source platform that can be assembled independently by different labs \cite{robel2019, replab2019, trifinger}. A third approach considers benchmarking performance on large offline datasets such as robot grasps on 3D object models, e.g., EGAD \cite{EGAD} and Dex-Net \cite{mahler2017dexnet}; RGBD scans and meshes of real-world common household objects, e.g., the YCB Object and Model set \cite{ycb}; and video frames of robot experience, e.g., RoboNet \cite{dasari2020robonet}. These datasets have been used to explore and compare algorithms \cite{kim2021simulation}, but they limit evaluation to states within the dataset.
%, but they require future states visited by the robot to belong to the distribution of the offline dataset, which generally does not hold for robots that actively influence the environment.

\textbf{Autonomous Fabric Folding:} Autonomous fabric manipulation is an active challenge in robotics. Maitin-Shepard et al. \cite{maitin2010cloth} and Doumanoglou et al. \cite{completepipeline} present early approaches to reliably fold towels and garments, respectively, from crumpled initial configurations. Weng et al. \cite{wengfabricflownet} and Ha et al. \cite{ha2021flingbot} develop learning-based algorithms for fabric manipulation using optical flow and dynamic flinging motions, respectively. However, these approaches were evaluated on dual-armed robots, which require coordination and are more costly. There has been significant recent interest in learning algorithms for unimanual (single-arm) fabric manipulation~\cite{fabric_vsf, seita_fabrics_2020, adi_descriptors}. These achieve strong results on fabric smoothing and folding tasks, but robust and precise unimanual T-shirt folding remains an open problem. Lin et al. \cite{softgym} propose an environment for fabric manipulation and benchmark several learning algorithms, but limit results to simulation. Garcia-Camacho et al. \cite{bimanual_benchmark} propose tasks for benchmarking bimanual cloth manipulation but allow the robot hardware to vary and do not evaluate learning algorithms.
\section{The Google Reach Testbed}
\label{sec:reach}

In this section, we review the most salient details of the Google Cloud Robotics testbed \cite{reach2022pyreach} as it relates to this case study.

\subsection{Hardware}\label{ssec:hardware}

See Figure~\ref{fig:reach-cell} for an image of the workcell. The robot is a single Universal Robot UR5e arm equipped with a Piab piSOFTGRIP vacuum-driven soft 3-jaw gripper \cite{piab}. The workcell is equipped with 4 Intel Realsense D415 cameras which each capture 640 $\times$ 360 RGB images at 20 FPS and 640 $\times$ 360 depth images at 1 FPS. The worksurface is a bright pink silicone mat and the garment is a blue crew-neck short sleeve T-shirt. The workcell is maintained by lab technicians who are onsite 8 hours a day to reset the robot and troubleshoot.

\subsection{Software} \label{ssec:software}

\begin{figure}
    \centering
    \includegraphics[width=0.5\textwidth]{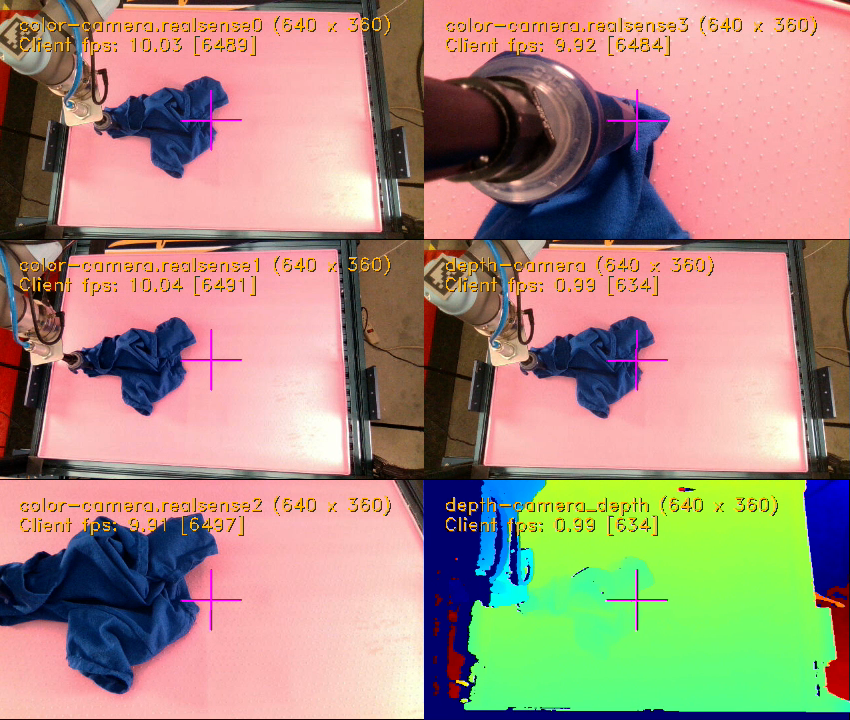}
    \caption{The client PyReach viewer, which updates the RGB images from the workcell cameras at about 10 Hz and depth images at 1 Hz. Our algorithms use the overhead RGB images (top left panel).}
    \label{fig:reach-viewer}
\end{figure}

Reach includes PyReach, an open source Python library developed by Robotics at Google for interfacing with the Reach system. The software includes infrastructure for authenticated users to establish a network connection with the robot server over the Internet, a viewer tool for locally displaying the 4 workcell camera feeds in real time (Figure~\ref{fig:reach-viewer}), a simulated workcell that mimics the real workcell for safely testing motions prior to deployment on the real system, and utility functions such as a pixel-to-world transform using the depth camera and conversions between different pose representations.

PyReach also includes PyReach Gym, an application programming interface (API) modeled after OpenAI Gym \cite{openai_gym}. Remote agents receive observations of the environment and request actions through this interface. In particular, at each time step with frequency up to 10 Hz, a remote agent can receive the joint angles and Cartesian pose of the arm, the binary state of the gripper (closed or open), and camera observations. The agent specifies an action to execute as a desired pose of the arm in either joint or Cartesian space and a desired binary state of the gripper.

\subsection{Garment Folding Case Study: Problem Definition}\label{ssec:ps}

\begin{figure}
    \centering
    \includegraphics[width=0.5\textwidth]{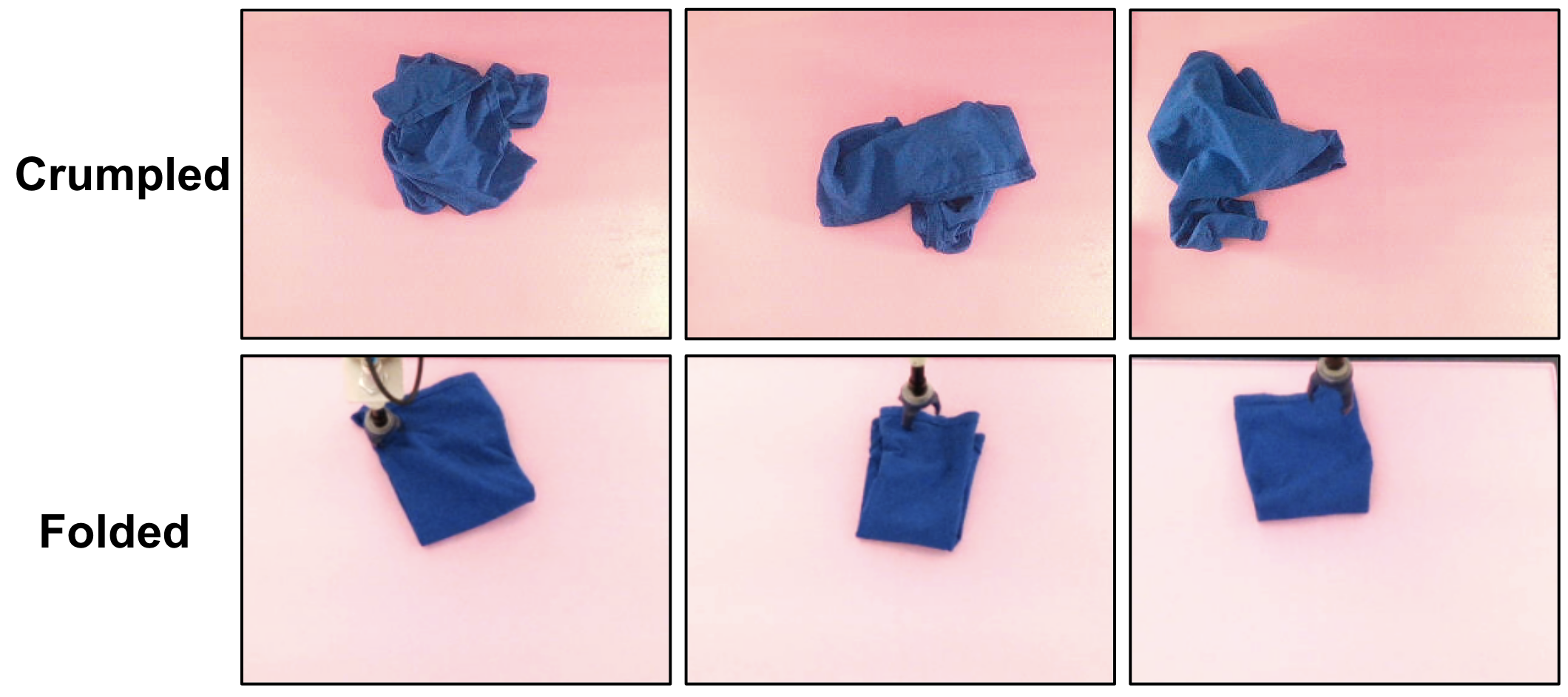}
    \caption{Examples of crumpled states (Row 1) and folded states (Row 2).}
    \label{fig:folded}
\end{figure}

We assume that the target folded configuration is known beforehand, that training and evaluation are performed in real (not simulation), that the hardware setup is as specified in Section~\ref{ssec:hardware}, and that the garment stays the same during training and evaluation.  %We also assume that target poses must be sent individually over the network instead of as a trajectory, which prevents highly dynamic scripted motions.
The task is to iteratively execute two procedures in a loop: (1) crumple the T-shirt and (2) fold the T-shirt. %Since automated classification of the state of the T-shirt is challenging, upon completing an episode, the images of the initial and final state are sent to a panel of 5 human judges to be labeled asynchronously through crowd computing. The verdict is then determined by the majority decision (3 or more out of 5); see Figure~\ref{fig:folded} for examples of successful and unsuccessful crumpling and folding. 
Crumpling is performed via a series of 6 random drops of the T-shirt resulting in an average of 37.5\% coverage (Section~\ref{ssec:smetrics}), where coverage is the fraction of the maximum 2D area the T-shirt is able to attain. %A crumpling episode is successful if the final state of the T-shirt is not easily discernible to a human judge. A folding episode is successful if the initial state is considered crumpled and the final state is neatly folded into the configuration in Figure~\ref{fig:folded}. Since crumpling is significantly easier than folding (e.g., via executing random actions), in this work we focus on the folding task. 
The folding task is to manipulate the T-shirt toward the target configuration in Figure~\ref{fig:folded}. We decompose the folding task into two subtasks: (1) \textit{flattening}, i.e., spreading out from an initially crumpled configuration until the garment is smooth, followed by (2) \textit{folding}, i.e., folding the t-shirt from initially flattened until sufficiently close to the target configuration. We measure folding accuracy with a combination of Intersection over Union (IOU) and wrinkle detection (Section~\ref{ssec:fmetrics}).

\section{Garment Folding Algorithms}
\label{sec:alg}

\setlength\extrarowheight{2pt}

Due to the unique challenges of the flattening and folding subtasks, we benchmark each subtask with its own set of algorithms. Hyperparameter and implementation details for all algorithms are available in the appendix, and notation for this section is defined in Table~\ref{tab:def}. With the exception of Section~\ref{ssec:dm}, all actions are quasistatic pick-and-place actions from pick point $p_0$ to place point $p_1$, where $p_0$ and $p_1$ are specified as $(x,y)$ coordinates in pixel space; see Section~\ref{ssec:esetup} for implementation details.

{\parindent0pt
\begin{table}[]
    \centering
    \begin{tabular}{|c|p{6.75cm}|}
        \hline
        $\mathbf{o}_t$ & Cropped RGB observation of the workcell state from the overhead Realsense camera at timestep $t$ (Figure~\ref{fig:reach-viewer}).\\ \hline
        $\mathbf{a}_t$ & The action at time $t$, expressed as a pick-and-place action $(p_0, p_1)$ in pixel coordinates except in Section~\ref{ssec:dm}. \\ \hline
        $\mathbf{o}_t^m$ & Color-thresholded mask of the T-shirt analytically computed from $\mathbf{o}_t$. \\
        \hline
        $\textrm{com}(\mathbf{o}_t^m)$ & A function that returns the \textit{visual} center of the T-shirt. \\ \hline
        $\textrm{cover}(\mathbf{o}_t^m)$ & A function that computes the 2D fabric coverage. \\ \hline
        $\mathcal{T}$ & A template image of a fully flattened shirt in the workspace. \\ \hline
        %$p_0$ & The initial point of a pick-and-place action in pixel space, i.e. the $(x,y)$ coordinates of the pick.\\
        %\hline
        %$p_1$ & The final point of a pick-and-place action in pixel space, i.e. the $(x,y)$ coordinates of the place.\\
        %\hline
        
    \end{tabular}
    \caption{Notation for Section~\ref{sec:alg}.}
    \label{tab:def}
\end{table}
}

\subsection{Flattening: 4 New Algorithms} 

\subsubsection{Learned Pick-Analytic Place (LP$_0$AP$_1$)}\label{ssec:il}

\begin{figure}
    \centering
    \includegraphics[width=0.5\textwidth]{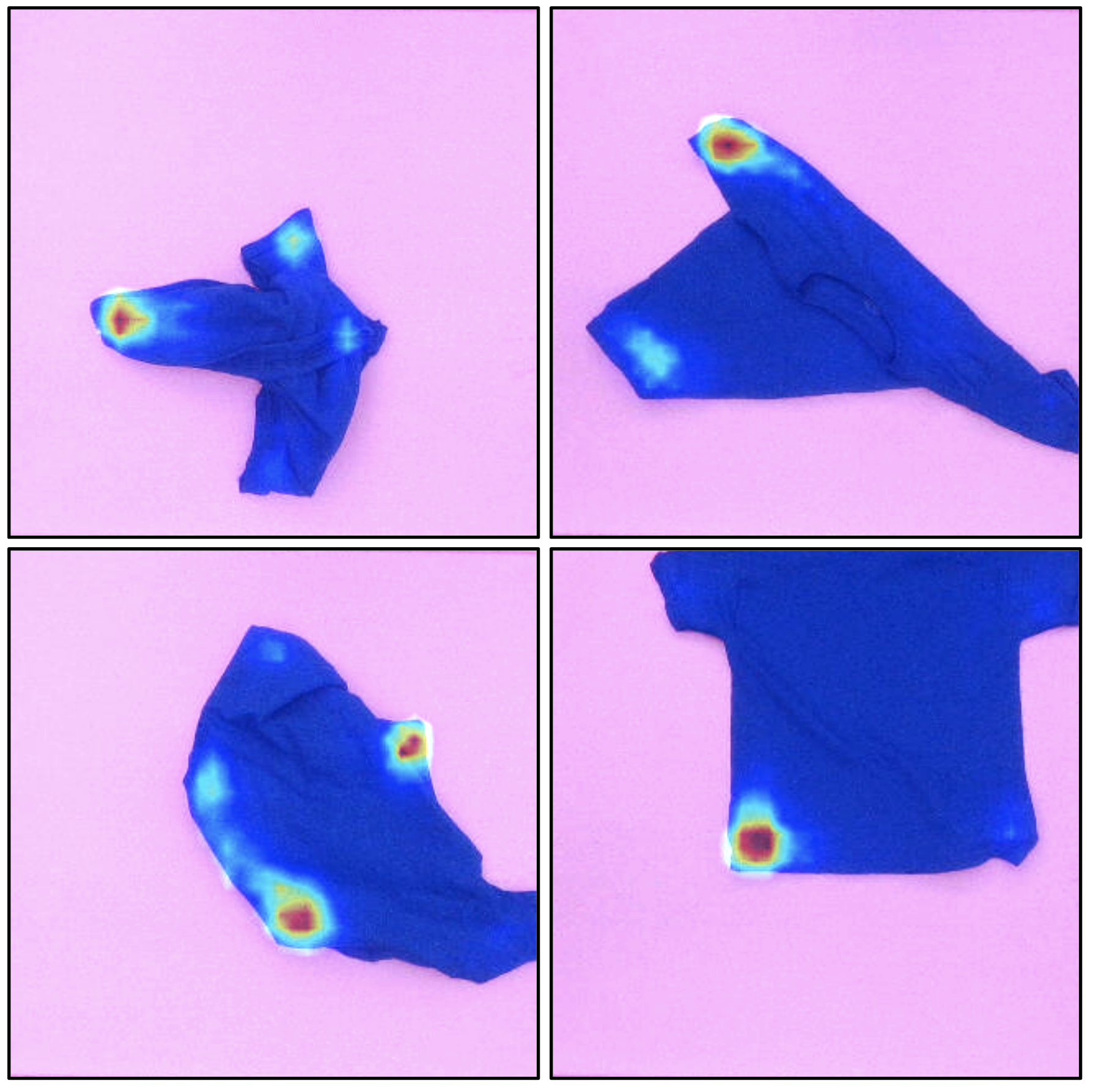}
    \caption{LP$_0$AP$_1$ pick point predictions on the test set. Bright red and yellow regions correspond to high probability pick points. The output heatmap is able to capture the multimodality in human actions.}
    \label{fig:il-heatmap}
\end{figure}

Inspired by prior work in imitation learning for fabric manipulation \cite{seita_fabrics_2020, lazydagger}, we develop an algorithm to learn pick points from human demonstrations. Since we empirically observe that human-selected pick points combined with analytic placing performs well on flattening, we propose only learning the pick points $p_0$ and analytically computing place points with the strategy in Section~\ref{ssec:analytic} to improve sample efficiency. While other work has considered learning a pick-conditioned place policy for fabric manipulation \cite{wu2020learning}, we define analytic placing actions that make the pick-conditioned policy unnecessary. To handle the inherent multimodality in the human policy, we train a fully convolutional network (FCN) \cite{FCNs} to output heatmaps corresponding to probability density instead of regressing to individual actions (Figure~\ref{fig:il-heatmap}). The FCN can be interpreted as an implicit energy-based model \cite{Florence2021ImplicitBC, Zeng2020TransporterNR} where the state and action pairs are the receptive fields of the network. As in DAgger \cite{dagger}, we reduce distribution shift by iteratively adding on-policy action labels to the dataset.

\subsubsection{Learning Keypoints (KP)}\label{ssec:kp}

\begin{figure}
    \centering
    \includegraphics[width=0.5\textwidth]{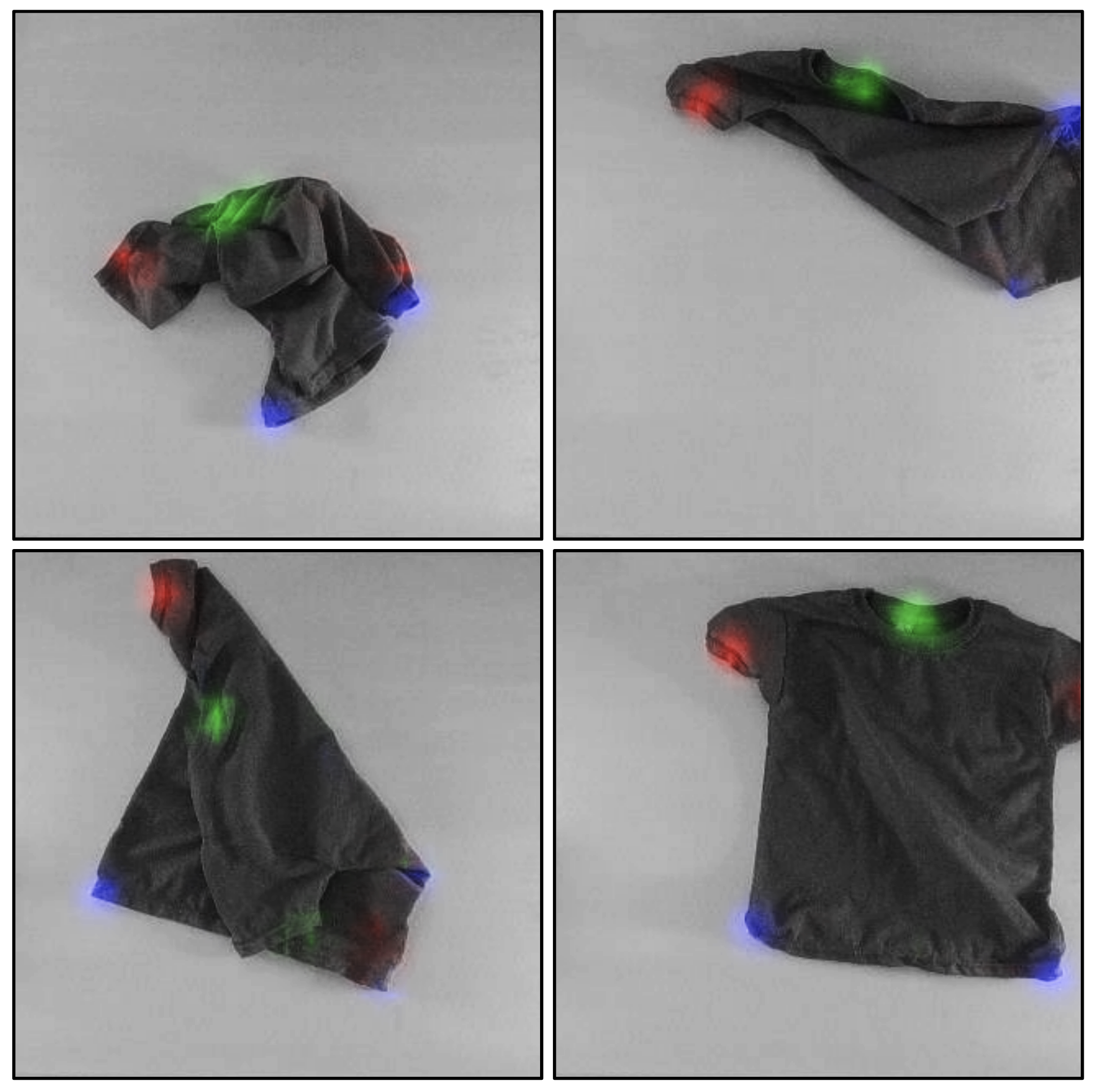}
    \caption{KP predictions on the test set. The predicted collar is colored green, the two sleeves are red, and the two base points are blue. Shirt images are shown in grayscale for viewing convenience.}
    \label{fig:kpts}
\end{figure}

This approach separates perception from planning and proposes to only learn the perception component. Specifically, we collect a hand-labeled dataset of images with up to 5 visible keypoints on the fabric corresponding to the collar, 2 sleeves, and 2 base corners (Figure~\ref{fig:kpts}). While the dataset generation policy is open-ended for this approach, we choose to first train an initial KP policy on random data (Section~\ref{ssec:rand}) and then augment the dataset with states encountered under the policy to mitigate distribution mismatch similar to DAgger \cite{dagger}. We train a FCN with 3 output heatmaps to predict each of the 3 classes of keypoints separately. Using keypoint predictions, we propose an analytic corner-pulling policy inspired by \cite{seita_fabrics_2020} that iteratively moves the keypoints from their current positions to their respective locations on a template flattened shirt $\mathcal{T}$. To reduce ambiguity, we compute the rotation and translation of $\mathcal{T}$ that best matches the current state and first move the keypoint farthest from its target location to its destination. To our knowledge, the combination of the FCN for multi-class keypoint prediction, T-shirt template matching, and corner pulling is a novel flattening policy.

\subsubsection{Coverage Reward Learning (CRL)}\label{ssec:fdyn}

This approach seeks to learn a reward function corresponding to fabric coverage $\textrm{cover}(\cdot)$ from data and execute a policy using this reward function. We learn this reward with self-supervised learning and execute a greedy policy that seeks to maximize the 1-step reward at each time step. Specifically, we fit a Convolutional Neural Network (CNN) $R_\theta(\mathbf{o}_t, \mathbf{a}_t$) to the scalar change in coverage (i.e., $\textrm{cover}(\mathbf{o}_{t+1}^m) - \textrm{cover}(\mathbf{o}_t^m)$) that results from executing action $\mathbf{a}_t$ on $\mathbf{o}_t$. At execution time we randomly sample thousands of pick points on the fabric mask $\mathbf{o}_t^m$ and place points in the workspace and select the action with the highest predicted change in coverage. To our knowledge, greedy planning over a learned model of coverage dynamics for fabric flattening is novel. Once again, the dataset generation policy is a design choice; here, we opt for a random action policy (Section~\ref{ssec:rand}) to enable large-scale self-supervised data collection and increase data diversity.

%Here we seek to first learn a model of the visual dynamics of the T-shirt from self-supervised data and then plan over this learned model to extract a policy. A forward visual dynamics model $g(\mathbf{o}_t, \mathbf{a}_t$) learns to predict the image observation $\mathbf{o}_{t+1}$ that results from executing action $\mathbf{a}_t$ on $\mathbf{o}_t$. Planning over this model requires an action sampling strategy such as random sampling or the cross entropy method \todo{cite} to find the best action according to a cost function evaluated on $\mathbf{o}_{t+1}$, such as pixel coverage $\textrm{cover}(\cdot)$ for flattening. Since we intend to use the model for flattening, we can simplify the network architecture and learning problem by modeling \textit{coverage} dynamics instead of full pixel dynamics. To our knowledge, planning over coverage dynamics instead of full pixel dynamics is novel. Specifically, we use Reach to collect a large self-supervised dataset generated by a random action policy (Section~\ref{ssec:rand}). We then train a Convolutional Neural Network (CNN) $g_\theta(\mathbf{o}_t, \mathbf{a}_t$) to output the scalar change in coverage resulting from this action, i.e., $\textrm{cover}(\mathbf{o}_{t+1}^m) - \textrm{cover}(\mathbf{o}_t^m)$. At execution time we  \todo{describe/justify dataset choice, maybe update to IRL-Greedy}

\subsubsection{Drop (DROP)}\label{ssec:dm}

Inspired by Ha et al. \cite{ha2021flingbot}, we investigate whether dynamic motions can leverage aerodynamic effects to accelerate the flattening of the shirt when combined with Approach~\ref{ssec:il}. %Since highly dynamic trajectories are currently infeasible on Reach (Section~\ref{ssec:ps}), 
We propose a simple vertical drop primitive that grabs the visual center of mass $\textrm{com}(\mathbf{o}_t^m)$, lifts the shirt into the air, and releases. We profile the coverage dynamics of the drop and the LP$_0$AP$_1$ pick-and-place and run Q-value iteration to determine which primitive to execute (i.e., drop or pick-and-place) given a discretized version of the current coverage $\textrm{cover}(\mathbf{o}_t^m)$. Q-value iteration on the following reward function produces a policy that minimizes the total number of actions required to flatten the shirt:

% \todo{@Ryan we should decide if we're putting the reward function or coverage here, if reward function, need to replace T with 1. If describing the extracted policy, we can put T and say drop vs. pickplace}
$$r(s = \textrm{cover}(\cdot), a) = 
    \begin{cases}
      -1 & \textrm{cover}(\cdot) < C \\
      0 & \textrm{cover}(\cdot) \geq C \\
   \end{cases}
$$
where $C$ is a coverage threshold defined in Section~\ref{ssec:smetrics} and cover($\cdot$) is the discretized current coverage. %To our knowledge, model-based tabular RL with garment manipulation primitives as the action space is novel.

\subsection{Flattening: 4 Baselines}

\subsubsection{Random (RAND)}\label{ssec:rand}

As a simple baseline, we implement a random pick-and-place policy that selects $p_0$ uniformly at random from $\mathbf{o}_t^m$ and $p_1$ uniformly at random in the workspace within a maximum distance from $p_0$.

\subsubsection{Human Teleoperation (HUMAN)}\label{ssec:humans}

As an upper bound on performance and action efficiency, a human selects pick and place points through a point-and-click interface (see the appendix for details).

\subsubsection{Analytic Edge-Pull (AEP)}\label{ssec:analytic}

We implement a fully analytic policy to explore to what extent learning is required for the T-shirt flattening task. The policy seeks to flatten the shirt by picking the edges and corners and pulling outwards. Formally, we sample $p_0$ uniformly from the set of points in the shirt mask $\mathbf{o}_t^m$ that are within a distance $k$ from the perimeter of $\mathbf{o}_t^m$, where $k$ is a hyperparameter. Given $p_0$, we compute $p_1$ by pulling a fixed distance $l$ in the direction of the average of two unit vectors: (1) away from the visual center of mass $\textrm{com}(\mathbf{o}_t^m)$ and (2) toward the nearest pixel outside $\mathbf{o}_t^m$.

\subsubsection{Learning an Inverse Dynamics Model (IDYN)}\label{ssec:idyn}

A inverse dynamics model $f(\mathbf{o}_t, \mathbf{o}_{t+1})$ produces the action $\mathbf{a}_{t}$ that causes the input transition from $\mathbf{o}_t$ to $\mathbf{o}_{t+1}$. Here we implement the algorithm proposed by Nair et al. \cite{Nair2017CombiningSL}, which learns to model visual inverse dynamics. Specifically, we approximate the dynamics with a Siamese CNN $f_\theta(\cdot,\cdot)$ trained on the random action dataset collected in Section~\ref{ssec:fdyn}. As in \cite{Nair2017CombiningSL}, the network factors the action by predicting the pick point $p_0$ before the pick-conditioned place point $p_1$ to improve sample efficiency. During policy evaluation, the inputs to the network are the current observation $\mathbf{o}_t$ and the template goal observation $\mathcal{T}$.

%\subsubsection{Transporter Networks (TNET)}\label{ssec:tnet}

%We implement Transporter Networks proposed by Zeng et al. \cite{Zeng2020TransporterNR}, a model architecture designed specifically for spatial displacement. \todo{@Adi add more details here}

\subsection{Folding Algorithms}

\subsubsection{Human Teleoperation (HUMAN)}\label{ssec:humanf}

As an upper bound on performance, a human chooses pick and place points for folding through a point-and-click interface.

\subsubsection{Analytic Shape-Matching (ASM)} \label{ssec:ASM}

Since the folding subtask is significantly more well-defined than flattening, we investigate whether an open-loop policy computed via shape matching can successfully fold the shirt. We specify a fixed sequence of folding actions with a single human demonstration. During evaluation, we compute rotations and translations of the corresponding template images to find the best match with $\mathbf{o}_t$ and transform the folding actions in the demonstration accordingly.

%\subsubsection{Learning Keypoints (KP)} This approach leverages the same perception network trained in Section~\ref{ssec:kp} but executes a different analytic policy. Instead of corner pulling, we execute a folding policy with the keypoints. Specifically, we move one base point to the other, move the corresponding sleeve to the other sleeve, move both sleeves toward the center of mass, and finally move the collar toward the base points. \todo{modify as appropriate}

\subsubsection{Learned Pick-Learned Place (LP$_0$LP$_1$)} This approach is identical to Section~\ref{ssec:il} but learns both pick points and place points, as the analytically computed place point is designed for flattening. Since folding demonstrations are difficult to obtain (the garment must be flattened first) and successful folding episodes are short-horizon and visually similar, we collect only two demonstrations and augment the data by a factor of 20 with affine transforms that encourage rotational and translational invariance.

\subsubsection{Fully Autonomous Flattening with Analytic Shape-Matching (A-ASM)} The algorithms above are evaluated after the garment is fully flattened via human teleoperation to study the folding subtask in isolation. This approach combines the best performing autonomous flattening algorithm (i.e., LP$_0$AP$_1$) with ASM (Section~\ref{ssec:ASM}) to evaluate the performance of a fully autonomous pipeline for manipulating the garment from crumpled to folded.

%\subsubsection{Learning an Inverse Dynamics Model (IDYN)} This approach uses the same inverse dynamics model trained in Section~\ref{ssec:idyn}. However, since folding is more well-defined than flattening, we can now provide a sequence of subgoal images during planning instead of the single template $\mathcal{T}$. This may improve the model's generalization, but it may be difficult to encounter these highly structured transitions in a randomly collected dataset.

%\subsubsection{Transporter Networks (TNET)}

%The approach is identical to Section~\ref{ssec:tnet} but we provide human demonstrations of folding instead of flattening.

%\todo{Add/subtract/modify algs as needed}
\section{Experiments}
\label{sec:exps}

\subsection{Experimental Setup}\label{ssec:esetup}
 
All actions executed on the robot are either a pick-and-place primitive $(p_0, p_1)$ or a drop primitive (for the DROP algorithm). During flattening, the pick-and-place primitive is a composition of 9 calls to the PyReach Gym API (Section~\ref{ssec:software}) that moves the gripper (oriented top-down) to $p_0$, lowers the gripper to grab the top layer of the fabric at $p_0$ (computed via a pixel-to-world transform using the depth camera), lifts the gripper, translates to $p_1$, and releases. During folding, the pick-and-place primitive moves the gripper more slowly; performs a deeper, possibly multi-layer pick; and lowers the gripper at $p_1$ instead of letting go in the air. The drop primitive uses a similar vertical pick to grab the shirt at $\textrm{com}(\mathbf{o}_t^m)$, raise it to a predefined height over the center of the workspace, and let go. After executing an action, the arm is commanded to clear the field of view of the camera to prevent occlusion in the image observation. See the appendix or code for exact implementation details. During data collection, actions are chosen either autonomously (e.g., with RAND in Section~\ref{ssec:rand}) or by a human via a point-and-click graphical user interface (see the appendix). At execution time, actions are specified by trained model outputs.
 
To improve the performance of the deployed flattening algorithms, we include two additional primitives: (1) a recentering primitive for when the shirt has drifted too far from the center of the workspace, and (2) a recovery primitive that executes a random action when the coverage is stalled for an extended period of time. See the appendix for ablation studies suggesting the usefulness of such primitives.

% We make the following critical modifications to improve the performance of our learned algorithms. Through ablations, we later demonstrate the efficacy of these simple additions in improving smoothing success. \todo{actually add the ablation results}
% \begin{enumerate}
%     \item Use a recentering primitive when the shirt is too far to the edge to improve future manipulability.
%     \item Use random recovery actions when coverage is sufficiently low, and the algorithm is not making progress in increasing coverage.
% \end{enumerate}
 
\subsection{Flattening Metrics}\label{ssec:smetrics}

We perform 10 trials of all flattening algorithms from an initially crumpled state (Figure~\ref{fig:folded}). Crumpling is performed autonomously via a series of 6 actions, each of which grabs the T-shirt at a random point, quickly lifts it into the air, and releases, resulting in an initial coverage of $37.5\% \pm 14.9\%$ over 45 trials. In Table~\ref{tab:flatten} we report maximum coverage as a percentage of the pixel coverage of a fully flattened shirt, i.e. 47,000 pixels in the shirt mask $\mathbf{o}_t^m$. We also report the number of samples used to train the algorithm, the execution time per action, and the number of actions executed, where we allow a maximum of 100 actions but terminate early if a coverage threshold is reached ($C =$ 45,000 pixels or 96\% of maximally flattened).

\subsection{Folding Metrics}\label{ssec:fmetrics}

We perform 5 trials of all folding algorithms from an initially flattened state. A-ASM initial states are flattened by LP$_0$AP$_1$ while all other initial states are flattened via human teleoperation. In Table~\ref{tab:fold} we report the number of actions and execution time per action, and we measure the quality of the final state against a goal configuration (Figure~\ref{fig:folded}) according to two metrics: (1) intersection over union (IoU) and (2) a penalty for edges and wrinkles. IoU is calculated between the shirt mask and the goal template, after rotating and translating the goal to best match the shirt mask. The wrinkle penalty calculates the fraction of pixels in the interior of the shirt mask detected as edges by the Canny edge detector \cite{canny1986computational}. A high-quality folding episode achieves a high IoU score and low edge penalty; for reference, the scores for a fully folded goal image are provided in Table~\ref{tab:fold} as GOAL.

\subsection{Flattening Results} 

\begin{figure}
    \centering
    \includegraphics[width=1.05\linewidth]{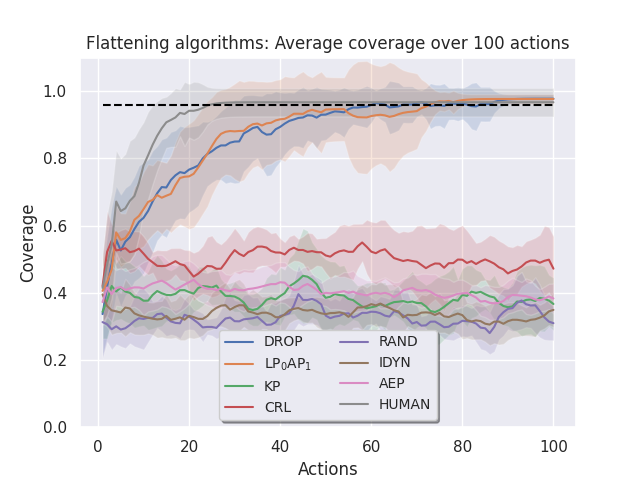}
    \caption{Coverage vs. time plot for the various flattening policies that we benchmark on the workcell, averaged across 10 rollouts. Shading represents one standard deviation, and the horizontal dashed line is the flattening success threshold (96\%).}
    \label{fig:avgcovvstime}
\end{figure}

See Table~\ref{tab:flatten} and Figure~\ref{fig:avgcovvstime} for results. We find that fully analytical policies such as RAND and AEP are unable to attain high coverage while HUMAN is able to consistently flatten the garment in 11.9 actions on average, suggesting the efficacy of the pick-and-place action primitive and the value of intelligently selecting pick points. Interestingly, we find that despite training an inverse dynamics model on nearly 4,000 real samples, IDYN is unable to outperform RAND. We hypothesize that the fully flattened goal image $\mathcal{T}$ provided as input is too distant from the encountered states, resulting in a data sample outside the training data distribution. While a more fine-grained sequence of subgoal images can mitigate this, such a sequence is not well-defined for flattening, suggesting IDYN is not well-suited for flattening without significant modifications. 

CRL is better able to leverage the large self-supervised dataset as it attains higher coverage, though it does require more time per action due to executing thousands of forward passes through the network during planning. However, since the dataset is generated by RAND, which achieves an average maximum coverage of only 55.0\%, CRL has trouble producing high-quality actions in the high coverage regime, where it has encountered relatively little data. Modifications to the dataset such as including demonstration data or actively interleaving data collection and training with policy execution could lead to further improvements and is an interesting direction for future work. KP is also able to improve upon RAND but struggles with achieving high coverage, despite having access to hand-labeled data relatively within the distribution of encountered states. While KP achieves a higher maximum coverage than AEP and RAND, it is prone to executing regressive actions that prevent it from maintaining this coverage. Results suggest that KP can be improved by (1) autonomous labeling, e.g. with fiducial markers, to avoid human error on challenging garment states with high self-occlusion, and (2) improvements to the analytic corner-pulling policy, which, for example, can struggle when all visible keypoints are positioned correctly but others are layered underneath the garment.

We find that LP$_0$AP$_1$ significantly outperforms all other algorithms, rivaling HUMAN-level performance by consistently reaching the threshold coverage $C$ in less than 3 times the amount of actions as HUMAN. We hypothesize that this is due to increased sample efficiency from analytic placing in conjunction with the modeling power of the FCN, which exhibits equivariance by sharing parameters for pixel predictions and is an implicit energy-based model like other state-of-the-art architectures \cite{Florence2021ImplicitBC, Zeng2020TransporterNR}. %We also run ablations and find that both recovery actions and shrinking actions in high coverage states help performance, without which LP$_0$AP$_1$ takes $52.2 \pm 28.9$ actions and $48.8 \pm 30.0$ actions to converge respectively. %This indicates that the simple practical modifications we work into our algorithm are crucial for improving upon naive behavior cloning. \todo{@Ryan check this.}

Finally, we find that DROP, which converges through Q-iteration to a policy that executes a drop if coverage is below $45\%$ and LP$_0$AP$_1$ otherwise, is unable to improve upon LP$_0$AP$_1$. This may occur due to our modeling of the coverage dynamics of LP$_0$AP$_1$ as the same regardless of the current coverage, whereas in reality, LP$_0$AP$_1$ improves coverage faster in lower-coverage states (Figure \ref{fig:avgcovvstime}).
% We observe that the drop primitive reduces to a geometric series preceding LP$_0$AP$_1$, which fails to improve LP$_0$AP$_1$ since it is already able to quickly exit the low coverage regime (figure \ref{fig:avgcovvstime}). 
Nevertheless, the DROP framework may be an effective way to combine multiple action primitives given more powerful dynamic primitives, such as bimanual actions that can better leverage aerodynamic effects \cite{ha2021flingbot}.

\begin{table}[t]
\caption{
\small
Flattening results. We report maximum coverage, number of actions, number of samples in the dataset, and evaluation time, where averages and standard deviations are computed over 10 trials.
}
\centering
\begin{tabular}{l | l l r r}\label{tab:flatten}
\textbf{Algorithm} & \textbf{\% Coverage} & \textbf{Actions} & \textbf{Dataset} & \textbf{Time/Act (s)} \\ \hline 
RAND & 55.0 $\pm$ 6.0 & 100.0 $\pm$ 0.0 & N/A & 23.9 $\pm$ 2.5 \\
HUMAN & \textbf{97.7 $\pm$ 3.9} & 11.9 $\pm$ 5.3 & N/A & 45.1 $\pm$ 18.6 \\
AEP & 55.3 $\pm$ 5.5 & 100.0 $\pm$ 0.0 & N/A & 24.6 $\pm$ 2.0  \\
IDYN & 57.0 $\pm$ 5.9 & 100.0 $\pm$ 0.0 & 3936 & 23.7 $\pm$ 3.7 \\
KP & 72.4 $\pm$ 9.2 & 100.0 $\pm$ 0.0 & 681 & 25.7 $\pm$ 2.7 \\
CRL & 73.8 $\pm$ 8.4 & 100.0 $\pm$ 0.0 & 3936 & 32.1 $\pm$ 5.3 \\
%TNET & TODO & TODO & TODO & TODO\\
DROP & \textbf{97.7 $\pm$ 1.3} & 38.6 $\pm$ 20.6 & 524 & 25.7 $\pm$ 0.8  \\
LP$_0$AP$_1$ & \textbf{97.7 $\pm$ 1.4} & 31.9 $\pm$ 17.2 & 524 & 25.6 $\pm$ 0.9  \\
\end{tabular}
%\vspace*{-10pt}
\label{tab:analytic}
\end{table}

\subsection{Folding Results}

\begin{figure}
    \centering
    \includegraphics[width=0.5\textwidth]{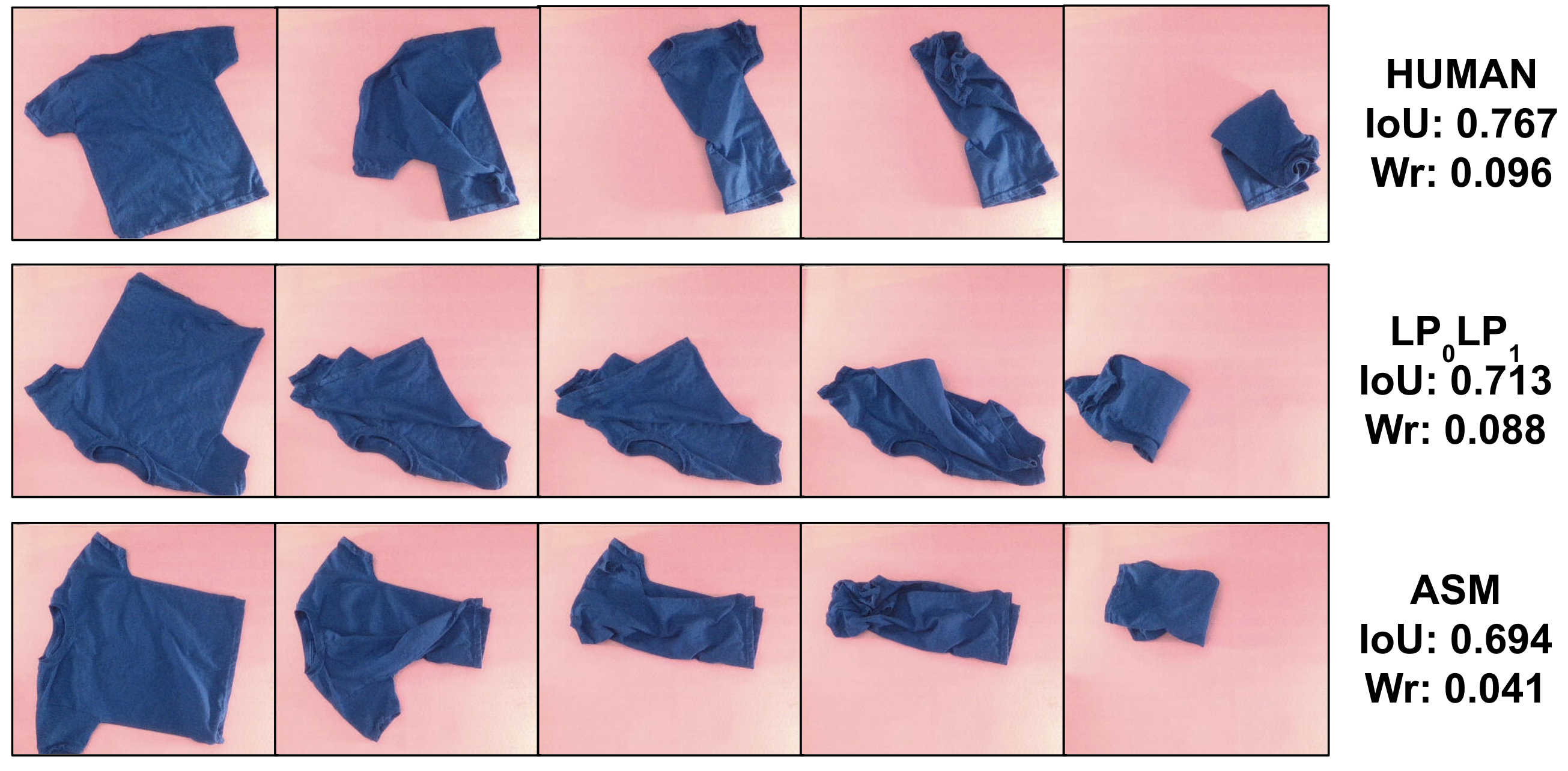}
    \caption{Representative episodes of the folding subtask executed by HUMAN (Row 1), LP$_0$LP$_1$ (Row 2), and ASM (Row 3). LP$_0$LP$_1$ and ASM achieve performance competitive with human teleoperation.}
    \label{fig:foldrollout}
\end{figure}

See Table~\ref{tab:fold} for results and Figure~\ref{fig:foldrollout} for folding episodes. The folding subtask presents unique challenges: (1) data collection and evaluation require an initially flattened state, which is difficult to attain through a remote interface, (2) slightly incorrect actions can dramatically alter the fabric state, often requiring re-flattening the garment, and (3) the single-arm pick-and-place primitive is not well-suited for the precise manipulation required for crisp garment folding. Indeed, we find that even with folding optimizations to the pick-and-place (Section~\ref{ssec:esetup}), a human teleoperator attains only 76\% of the goal IoU on average (Table~\ref{tab:fold}). However, we find that both ASM and LP$_0$LP$_1$ are able to effectively leverage the primitive to achieve near human-level performance, where ASM performs similarly to LP$_0$LP$_1$. We also find that the fully autonomous pipeline A-ASM is able to reach similar performance from an initially \textit{crumpled} state, setting a baseline score for the end-to-end folding task. Although ASM is open-loop and LP$_0$LP$_1$ learns from only 2 demonstrations, HUMAN cannot significantly outperform them due to the difficulty of correcting inaccurate actions in folding. Further progress on the folding subtask will likely benefit more from the design of manipulation primitives than from algorithmic innovations.

\begin{table}[t]
\caption{
\small
Folding results. We report intersection over union (IoU), wrinkle penalty, number of actions, and evaluation time, where averages and standard deviations are computed over 5 trials.
}
\centering
\begin{tabular}{l | l l r r}\label{tab:fold}
\textbf{Algo.} & \textbf{IoU ($\uparrow$)} & \textbf{Wrinkle ($\downarrow$)} & \textbf{Actions} & \textbf{Time/Act (s)} \\ \hline 
GOAL & 0.98 & 0.093 & N/A & N/A \\
HUMAN & \textbf{0.74 $\pm$ 0.06} & 0.088 $\pm$ 0.023 & 4.4 $\pm$ 0.5 & 63.8 $\pm$ 14.9 \\
ASM & \textbf{0.69 $\pm$ 0.08} & \textbf{0.087 $\pm$ 0.038} & 4.0 $\pm$ 0.0 & 35.1 $\pm$ 1.9  \\
%KP & TODO & TODO & TODO & TODO \\
LP$_0$LP$_1$ & 0.68 $\pm$ 0.08 & 0.112 $\pm$ 0.032 & 4.0 $\pm$ 0.0 & 35.7 $\pm$ 1.3  \\
A-ASM & 0.62 $\pm$ 0.12 & 0.112 $\pm$ 0.038 & 4.0 $\pm$ 0.0 & 35.5 $\pm$ 1.7 \\
%IDYN & TODO & TODO & TODO & TODO \\
%TNET & TODO & TODO & TODO & TODO\\
\end{tabular}
%\vspace*{-10pt}
\label{tab:analytic}
\end{table}

%\vspace{-0.18in}
\section{Conclusion and Future Work}
\label{sec:discussion}
In this work, we benchmark novel and existing algorithms for T-shirt smoothing and folding tasks. We find that policies that combine learning with analytical methods achieve the highest performance in practice, suggesting the value of future work in this area.

%Through our experiments on the Reach testbed, we find at the moment that the highest performing algorithm on fabric smoothing tasks is imitation learning (LP$_0$AP$_1$), which effectively and rapidly smooths the fabric, as well as analytic folding, which executes a folding routine with information from just a single rollout of folding from a supervisor, an ability that is useful in practice. A-ASM achieves an end-to-end folding performance, obtaining $84\%$ of human IoU while having only $27\%$ more wrinkles than humans.

%We find significant differences in performance among our algorithms, which are built on top of prior work in the field, as well as algorithms studied for deformable manipulation. These discrepancies in performance arise from a variety of factors, including but not limited to: 1) balancing how much needs to be learned versus computed analytically, and 2) how well these algorithms remain within their training data distributions. We also find that simple yet highly practical primitives such as recovery actions significantly improve performance. 

% We do \todo{insert}. Results suggest \todo{insert - empirical takeaways into what worked and what didn't work}. %Use of movement primitives such as the pick-and-place are well-suited for applications with highly variable latency such as a cloud robotics testbed, as they simplify human teleoperation and make the learning problem more tractable.

Remote robotics research poses both opportunities and challenges. On the one hand, the ability to access a robot from anywhere, the abstraction of robot operations behind an intuitive API, the setup and maintenance by dedicated staff, and the consistency of the task environment all contributed to quick and convenient experimentation. On the other hand, onsite technicians have limited availability, variable-latency 2D camera projections are at times insufficient for fully understanding the scene, and manual resets (e.g., flattening the T-shirt) become difficult to perform, suggesting the importance of learning self-supervised reset policies \cite{gupta2021resetfree} or integrating flattening more closely with folding.  

In future work, we will (1) further optimize performance on the unimanual folding task, (2) evaluate alternative approaches such as continuous control, reinforcement learning, and different action primitives, and (3) evaluate each algorithm's ability to generalize to other garments with variation in color, shape, size, and material. %We hope that this case study will facilitate reproducibility and accessibility by encouraging adoption of cloud-based robotics testbeds. 

\bibliographystyle{IEEEtran}
% \clearpage
\bibliography{references}

\begin{thebibliography}{10}
\providecommand{\url}[1]{#1}
\csname url@rmstyle\endcsname
\providecommand{\newblock}{\relax}
\providecommand{\bibinfo}[2]{#2}
\providecommand\BIBentrySTDinterwordspacing{\spaceskip=0pt\relax}
\providecommand\BIBentryALTinterwordstretchfactor{4}
\providecommand\BIBentryALTinterwordspacing{\spaceskip=\fontdimen2\font plus
\BIBentryALTinterwordstretchfactor\fontdimen3\font minus
  \fontdimen4\font\relax}
\providecommand\BIBforeignlanguage[2]{{%
\expandafter\ifx\csname l@#1\endcsname\relax
\typeout{** WARNING: IEEEtran.bst: No hyphenation pattern has been}%
\typeout{** loaded for the language `#1'. Using the pattern for}%
\typeout{** the default language instead.}%
\else
\language=\csname l@#1\endcsname
\fi
#2}}

\bibitem{imagenet}
J.~Deng, W.~Dong, R.~Socher, L.-J. Li, K.~Li, and L.~Fei-Fei, ``Imagenet: A
  large-scale hierarchical image database,'' in \emph{2009 IEEE Conference on
  Computer Vision and Pattern Recognition}, 2009, pp. 248--255.

\bibitem{ALFRED}
M.~Shridhar, J.~Thomason, D.~Gordon, Y.~Bisk, W.~Han, R.~Mottaghi,
  L.~Zettlemoyer, and D.~Fox, ``Alfred: A benchmark for interpreting grounded
  instructions for everyday tasks,'' \emph{2020 IEEE/CVF Conference on Computer
  Vision and Pattern Recognition (CVPR)}, pp. 10\,737--10\,746, 2020.

\bibitem{openai_gym}
G.~Brockman, V.~Cheung, L.~Pettersson, J.~Schneider, J.~Schulman, J.~Tang, and
  W.~Zaremba, ``Openai gym,'' \emph{arXiv preprint arXiv:1606.01540}, 2016.

\bibitem{softgym}
X.~Lin, Y.~Wang, J.~Olkin, and D.~Held, ``Softgym: Benchmarking deep
  reinforcement learning for deformable object manipulation,'' \emph{ArXiv
  preprint arXiv:2011.07215}, 2020.

\bibitem{Collins2019QuantifyingTR}
J.~J. Collins, D.~Howard, and J.~Leitner, ``Quantifying the reality gap in
  robotic manipulation tasks,'' \emph{2019 International Conference on Robotics
  and Automation (ICRA)}, pp. 6706--6712, 2019.

\bibitem{reach2022pyreach}
A.~Wong, A.~Zeng, A.~Bose, A.~Wahid, D.~Kalashnikov, I.~Krasin, J.~Varley,
  J.~Lee, J.~Tompson, M.~Attarian, P.~Florence, R.~Baruch, S.~Xu, S.~Welker,
  V.~Sindhwani, V.~Vanhoucke, and W.~Gramlich, ``Pyreach - python client sdk
  for robot remote control,'' \url{https://github.com/google-research/pyreach},
  2022.

\bibitem{piab}
``Piab pisoftgrip gripper,''
  \url{https://www.piab.com/en-us/suction-cups-and-soft-grippers/soft-grippers/pisoftgrip-vacuum-driven-soft-gripper-/pisoftgrip-/#overview},
  accessed: 2022-02-21.

\bibitem{seita_fabrics_2020}
D.~Seita, A.~Ganapathi, R.~Hoque, M.~Hwang, E.~Cen, A.~K. Tanwani,
  A.~Balakrishna, B.~Thananjeyan, J.~Ichnowski, N.~Jamali, K.~Yamane, S.~Iba,
  J.~Canny, and K.~Goldberg, ``{Deep Imitation Learning of Sequential Fabric
  Smoothing From an Algorithmic Supervisor},'' in \emph{Proc. IEEE/RSJ Int.
  Conf. on Intelligent Robots and Systems (IROS)}, 2020.

\bibitem{robotcluster}
S.~Bauer, F.~Widmaier, M.~W{\"{u}}thrich, N.~Funk, J.~U.~D. Jesus, J.~Peters,
  J.~Watson, C.~Chen, K.~Srinivasan, J.~Zhang, J.~Zhang, M.~R. Walter,
  R.~Madan, C.~B. Schaff, T.~Maeda, T.~Yoneda, D.~Yarats, A.~Allshire, E.~K.
  Gordon, T.~Bhattacharjee, S.~S. Srinivasa, A.~Garg, A.~Buchholz, S.~Stark,
  T.~Steinbrenner, J.~Akpo, S.~Joshi, V.~Agrawal, and B.~Sch{\"{o}}lkopf, ``A
  robot cluster for reproducible research in dexterous manipulation,''
  \emph{arXiv preprint arXiv:2109.10957}, 2021.

\bibitem{trifinger}
M.~W{\"u}thrich, F.~Widmaier, F.~Grimminger, S.~Joshi, V.~Agrawal, B.~Hammoud,
  M.~Khadiv, M.~Bogdanovic, V.~Berenz, J.~Viereck, M.~Naveau, L.~Righetti,
  B.~Sch{\"o}lkopf, and S.~Bauer, ``Trifinger: An open-source robot for
  learning dexterity,'' in \emph{Conf. on Robot Learning (CoRL)}, 2020.

\bibitem{benchmarking2022}
\BIBentryALTinterwordspacing
N.~Funk, C.~Schaff, R.~Madan, T.~Yoneda, J.~U. De~Jesus, J.~Watson, E.~K.
  Gordon, F.~Widmaier, S.~Bauer, S.~S. Srinivasa, T.~Bhattacharjee, M.~R.
  Walter, and J.~Peters, ``Benchmarking structured policies and policy
  optimization for real-world dexterous object manipulation,'' \emph{IEEE
  Robotics and Automation Letters}, vol.~7, no.~1, p. 478–485, Jan 2022.
  [Online]. Available: \url{http://dx.doi.org/10.1109/LRA.2021.3129139}
\BIBentrySTDinterwordspacing

\bibitem{Chen2021DexterousMP}
C.~Chen, K.~P. Srinivasan, J.~O. Zhang, and J.~Zhang, ``Dexterous manipulation
  primitives for the real robot challenge,'' \emph{ArXiv preprint
  arXiv:2101.11597}, 2021.

\bibitem{yoneda2021grasp}
T.~Yoneda, C.~Schaff, T.~Maeda, and M.~Walter, ``Grasp and motion planning for
  dexterous manipulation for the real robot challenge,'' \emph{arXiv preprint
  arXiv:2101.02842}, 2021.

\bibitem{mccarthy2021solving}
R.~McCarthy, F.~R. Sanchez, Q.~Wang, D.~C. Bulens, K.~McGuinness, N.~O'Connor,
  and S.~J. Redmond, ``Solving the real robot challenge using deep
  reinforcement learning,'' \emph{arXiv preprint arXiv:2109.15233}, 2021.

\bibitem{allshire2021transferring}
A.~Allshire, M.~Mittal, V.~Lodaya, V.~Makoviychuk, D.~Makoviichuk, F.~Widmaier,
  M.~W{\"u}thrich, S.~Bauer, A.~Handa, and A.~Garg, ``Transferring dexterous
  manipulation from gpu simulation to a remote real-world trifinger,''
  \emph{arXiv preprint arXiv:2108.09779}, 2021.

\bibitem{robotarium}
D.~Pickem, P.~Glotfelter, L.~Wang, M.~Mote, A.~Ames, E.~Feron, and
  M.~Egerstedt, ``The robotarium: A remotely accessible swarm robotics research
  testbed,'' in \emph{2017 IEEE International Conference on Robotics and
  Automation (ICRA)}, 2017, pp. 1699--1706.

\bibitem{duckietown}
L.~Paull, J.~Tani, H.~Ahn, J.~Alonso-Mora, L.~Carlone, M.~Cap, Y.~F. Chen,
  C.~Choi, J.~Dusek, Y.~Fang, D.~Hoehener, S.-Y. Liu, M.~Novitzky, I.~F.
  Okuyama, J.~Pazis, G.~Rosman, V.~Varricchio, H.-C. Wang, D.~Yershov, H.~Zhao,
  M.~Benjamin, C.~Carr, M.~Zuber, S.~Karaman, E.~Frazzoli, D.~Del~Vecchio,
  D.~Rus, J.~How, J.~Leonard, and A.~Censi, ``Duckietown: An open, inexpensive
  and flexible platform for autonomy education and research,'' in \emph{2017
  IEEE International Conference on Robotics and Automation (ICRA)}, 2017, pp.
  1497--1504.

\bibitem{softactorcritic}
T.~Haarnoja, A.~Zhou, P.~Abbeel, and S.~Levine, ``Soft actor-critic: Off-policy
  maximum entropy deep reinforcement learning with a stochastic actor,'' in
  \emph{Proc. Int. Conf. on Machine Learning (ICML)}, 2018.

\bibitem{td3}
S.~Fujimoto, H.~van Hoof, and D.~Meger, ``Addressing function approximation
  error in actor-critic methods,'' in \emph{Proc. Int. Conf. on Machine
  Learning (ICML)}, 2018.

\bibitem{robel2019}
M.~Ahn, H.~Zhu, K.~Hartikainen, H.~Ponte, A.~Gupta, S.~Levine, and V.~Kumar,
  ``Robel: Robotics benchmarks for learning with low-cost robots,'' in
  \emph{Conf. on Robot Learning (CoRL)}, 2019.

\bibitem{replab2019}
B.~Yang, J.~Zhang, V.~Pong, S.~Levine, and D.~Jayaraman, ``Replab: A
  reproducible low-cost arm benchmark platform for robotic learning,'' in
  \emph{{Proc. {IEEE} Int. Conf. Robotics and Automation (ICRA)}}, 2019.

\bibitem{EGAD}
D.~Morrison, P.~Corke, and J.~Leitner, ``Egad! an evolved grasping analysis
  dataset for diversity and reproducibility in robotic manipulation,''
  \emph{IEEE Robotics and Automation Letters}, vol.~5, pp. 4368--4375, 2020.

\bibitem{mahler2017dexnet}
J.~Mahler, J.~Liang, S.~Niyaz, M.~Laskey, R.~Doan, X.~Liu, J.~A. Ojea, and
  K.~Goldberg, ``Dex-net 2.0: Deep learning to plan robust grasps with
  synthetic point clouds and analytic grasp metrics,'' in \emph{Proc. Robotics:
  Science and Systems (RSS)}, 2017.

\bibitem{ycb}
\BIBentryALTinterwordspacing
B.~Calli, A.~Walsman, A.~Singh, S.~Srinivasa, P.~Abbeel, and A.~M. Dollar,
  ``Benchmarking in manipulation research: Using the yale-cmu-berkeley object
  and model set,'' \emph{IEEE Robotics \& Automation Magazine}, vol.~22, no.~3,
  p. 36–52, Sep 2015. [Online]. Available:
  \url{http://dx.doi.org/10.1109/MRA.2015.2448951}
\BIBentrySTDinterwordspacing

\bibitem{dasari2020robonet}
S.~Dasari, F.~Ebert, S.~Tian, S.~Nair, B.~Bucher, K.~Schmeckpeper, S.~Singh,
  S.~Levine, and C.~Finn, ``Robonet: Large-scale multi-robot learning,'' in
  \emph{Conf. on Robot Learning (CoRL)}, 2019.

\bibitem{kim2021simulation}
C.~M. Kim, M.~Danielczuk, I.~Huang, and K.~Goldberg, ``Simulation of
  parallel-jaw grasping using incremental potential contact models,''
  \emph{arXiv preprint arXiv:2111.01391}, 2021.

\bibitem{maitin2010cloth}
J.~Maitin-Shepard, M.~Cusumano-Towner, J.~Lei, and P.~Abbeel, ``{Cloth Grasp
  Point Detection Based on Multiple-View Geometric Cues with Application to
  Robotic Towel Folding},'' in \emph{{Proc. {IEEE} Int. Conf. Robotics and
  Automation (ICRA)}}, 2010.

\bibitem{completepipeline}
A.~Doumanoglou, J.~Stria, G.~Peleka, I.~Mariolis, V.~Petrík, A.~Kargakos,
  L.~Wagner, V.~Hlaváč, T.-K. Kim, and S.~Malassiotis, ``Folding clothes
  autonomously: A complete pipeline,'' \emph{IEEE Transactions on Robotics},
  vol.~32, no.~6, pp. 1461--1478, 2016.

\bibitem{wengfabricflownet}
T.~Weng, S.~Bajracharya, Y.~Wang, K.~Agrawal, and D.~Held, ``Fabricflownet:
  Bimanual cloth manipulation with a flow-based policy,'' in \emph{Conf. on
  Robot Learning (CoRL)}, 2021.

\bibitem{ha2021flingbot}
H.~Ha and S.~Song, ``Flingbot: The unreasonable effectiveness of dynamic
  manipulation for cloth unfolding,'' in \emph{Conf. on Robot Learning (CoRL)},
  2021.

\bibitem{fabric_vsf}
R.~Hoque, D.~Seita, A.~Balakrishna, A.~Ganapathi, A.~Tanwani, N.~Jamali,
  K.~Yamane, S.~Iba, and K.~Goldberg, ``{VisuoSpatial Foresight for Multi-Step,
  Multi-Task Fabric Manipulation},'' in \emph{Proc. Robotics: Science and
  Systems (RSS)}, 2020.

\bibitem{adi_descriptors}
A.~Ganapathi, P.~Sundaresan, B.~Thananjeyan, A.~Balakrishna, D.~Seita,
  J.~Grannen, M.~Hwang, R.~Hoque, J.~E. Gonzalez, N.~Jamali, K.~Yamane, S.~Iba,
  and K.~Goldberg, ``Learning to smooth and fold real fabric using dense object
  descriptors trained on synthetic color images,'' in \emph{{Proc. {IEEE} Int.
  Conf. Robotics and Automation (ICRA)}}, 2020.

\bibitem{bimanual_benchmark}
I.~Garcia-Camacho, M.~Lippi, M.~C. Welle, H.~Yin, R.~Antonova, A.~Varava,
  J.~Borras, C.~Torras, A.~Marino, G.~Alenyà, and D.~Kragic, ``Benchmarking
  bimanual cloth manipulation,'' \emph{IEEE Robotics and Automation Letters},
  vol.~5, no.~2, pp. 1111--1118, 2020.

\bibitem{lazydagger}
R.~Hoque, A.~Balakrishna, C.~Putterman, M.~Luo, D.~S. Brown, D.~Seita,
  B.~Thananjeyan, E.~Novoseller, and K.~Goldberg, ``{LazyDAgger}: Reducing
  context switching in interactive imitation learning,'' in \emph{International
  Conference on Automation Sciences and Engineering (CASE)}, 2021.

\bibitem{wu2020learning}
Y.~Wu, W.~Yan, T.~Kurutach, L.~Pinto, and P.~Abbeel, ``Learning to manipulate
  deformable objects without demonstrations,'' in \emph{Proc. Robotics: Science
  and Systems (RSS)}, 2020.

\bibitem{FCNs}
E.~Shelhamer, J.~Long, and T.~Darrell, ``Fully convolutional networks for
  semantic segmentation,'' \emph{IEEE Transactions on Pattern Analysis and
  Machine Intelligence}, vol.~39, pp. 640--651, 2017.

\bibitem{Florence2021ImplicitBC}
P.~R. Florence, C.~Lynch, A.~Zeng, O.~Ramirez, A.~Wahid, L.~Downs, A.~S. Wong,
  J.~Lee, I.~Mordatch, and J.~Tompson, ``Implicit behavioral cloning,'' in
  \emph{Conf. on Robot Learning (CoRL)}, 2021.

\bibitem{Zeng2020TransporterNR}
A.~Zeng, P.~R. Florence, J.~Tompson, S.~Welker, J.~Chien, M.~Attarian,
  T.~Armstrong, I.~Krasin, D.~Duong, V.~Sindhwani, and J.~Lee, ``Transporter
  networks: Rearranging the visual world for robotic manipulation,'' in
  \emph{Conf. on Robot Learning (CoRL)}, 2020.

\bibitem{dagger}
S.~Ross, G.~J. Gordon, and J.~A. Bagnell, ``A reduction of imitation learning
  and structured prediction to no-regret online learning,'' in
  \emph{International Conference on Artificial Intelligence and Statistics
  (AISTATS)}, 2011.

\bibitem{Nair2017CombiningSL}
A.~Nair, D.~Chen, P.~Agrawal, P.~Isola, P.~Abbeel, J.~Malik, and S.~Levine,
  ``Combining self-supervised learning and imitation for vision-based rope
  manipulation,'' \emph{2017 IEEE International Conference on Robotics and
  Automation (ICRA)}, pp. 2146--2153, 2017.

\bibitem{canny1986computational}
J.~Canny, ``A computational approach to edge detection,'' \emph{IEEE
  Transactions on Pattern Analysis and Machine Intelligence}, no.~6, pp.
  679--698, 1986.

\bibitem{gupta2021resetfree}
A.~Gupta, J.~Yu, T.~Z. Zhao, V.~Kumar, A.~Rovinsky, K.~Xu, T.~Devlin, and
  S.~Levine, ``Reset-free reinforcement learning via multi-task learning:
  Learning dexterous manipulation behaviors without human intervention,''
  \emph{arXiv preprint arXiv:2104.11203}, 2021.

\end{thebibliography}

\clearpage
% \newpage
\section{Appendix}
In Appendix~\ref{ssec:algdetails} and \ref{ssec:algdetails2}, we provide implementation and hyperparameter details for all flattening and folding algorithms. In Appendix~\ref{ssec:gui}, we provide information about the graphical user interface used to collect human-labeled data. In Appendix~\ref{ssec:primitives}, we describe the implementation details of the action primitives we use in this work. Finally, in Appendix~\ref{ssec:ablations}, we provide the results of ablation studies suggesting the usefulness of various design choices.

\subsection{Flattening Algorithm Details}\label{ssec:algdetails}

\subsubsection{RAND} As described in the main text, this baseline simply selects the pick point $p_0$ uniformly at random from the garment mask. The place point $p_1$ is sampled uniformly at random from the workspace, but resampled if $p_1$ is separated from $p_0$ by more than 50\% of the workspace in either $x$ or $y$. This prevents excessively large action deltas, which has been shown to be useful in prior fabric manipulation work \cite{fabric_vsf}.

\subsubsection{HUMAN}\label{ssec:appHUMAN} As described in the main text, this baseline allows the human to freely specify both pick and place points via the graphical user interface in Section~\ref{ssec:gui}.

\subsubsection{AEP}\label{ssec:appAEP}
Our analytic smoothing policy is based on the observation that generally pulling outwards on the shirt increases its coverage over time. Specifically, we do the following:
\begin{itemize}
    \item Compute the edge mask of the current shirt by using the formula $\texttt{xor}(\texttt{erode}(\mathbf{o}_t^m))$, where $\texttt{xor}(\cdot)$ is pixelwise exclusive or and $\texttt{erode}(\cdot)$ is a function that removes a set number of pixels (in our case, 40) at the boundary of the input mask.
    \item Sample uniformly among the resulting pixels to choose $p_0$.
    \item Choose the place point analytically by moving away from the center of mass and towards the edge of the shirt. Specifically, we compute $p_1$ as 
    $$
    p_1 = p_0 + k_1 \cdot (p_0 - \textrm{com}(\mathbf{o}_t^m)) + k_2 \cdot  (\textrm{bgd}(\mathbf{o}_t^m, p_0) - p_0)
    $$
    where $\textrm{bgd}(\mathbf{o}_t^m, p_0)$ is the closest background (i.e., non-shirt) pixel to $p_0$ and $k_1 = k_2 = 23$ are tuned constants, but action magnitudes are shrunk to $k_1 = k_2 = 14$ for higher coverage states (at or equal to $75\%$) to improve stability and convergence to a flattened state.
\end{itemize}

\subsubsection{IDYN} We first collect a dataset of 4402 random actions (90\% train / 10\% test split) by allowing the robot to run the RAND algorithm autonomously. The garment is reset after every 100 actions (with the procedure in Section~\ref{ssec:primitives}). We then train a Siamese CNN on $(\mathbf{o}_t, \mathbf{a}_t, \mathbf{o}_{t+1})$ tuples to output the $\mathbf{a}_t$ that takes $\mathbf{o}_t$ to $\mathbf{o}_{t+1}$. The convolutional layers are a ResNet-34 backbone with rectified linear unit (ReLU) output activation and pretrained (but not frozen) weights, shared between the two heads of the network. Each head encodes the input image into a vector of dimension 1,000. The two vectors are concatenated and passed through a fully connected layer with output size 2 and sigmoid activation, which generates the pick point. The predicted pick point is then concatenated with the two vectors and passed through another fully connected layer with output size 2 and sigmoid activation, which generates the pick-conditioned place point. The CNN has a total of 21.8 million parameters.

\begin{figure}
    \centering
    \includegraphics[width=0.49\textwidth]{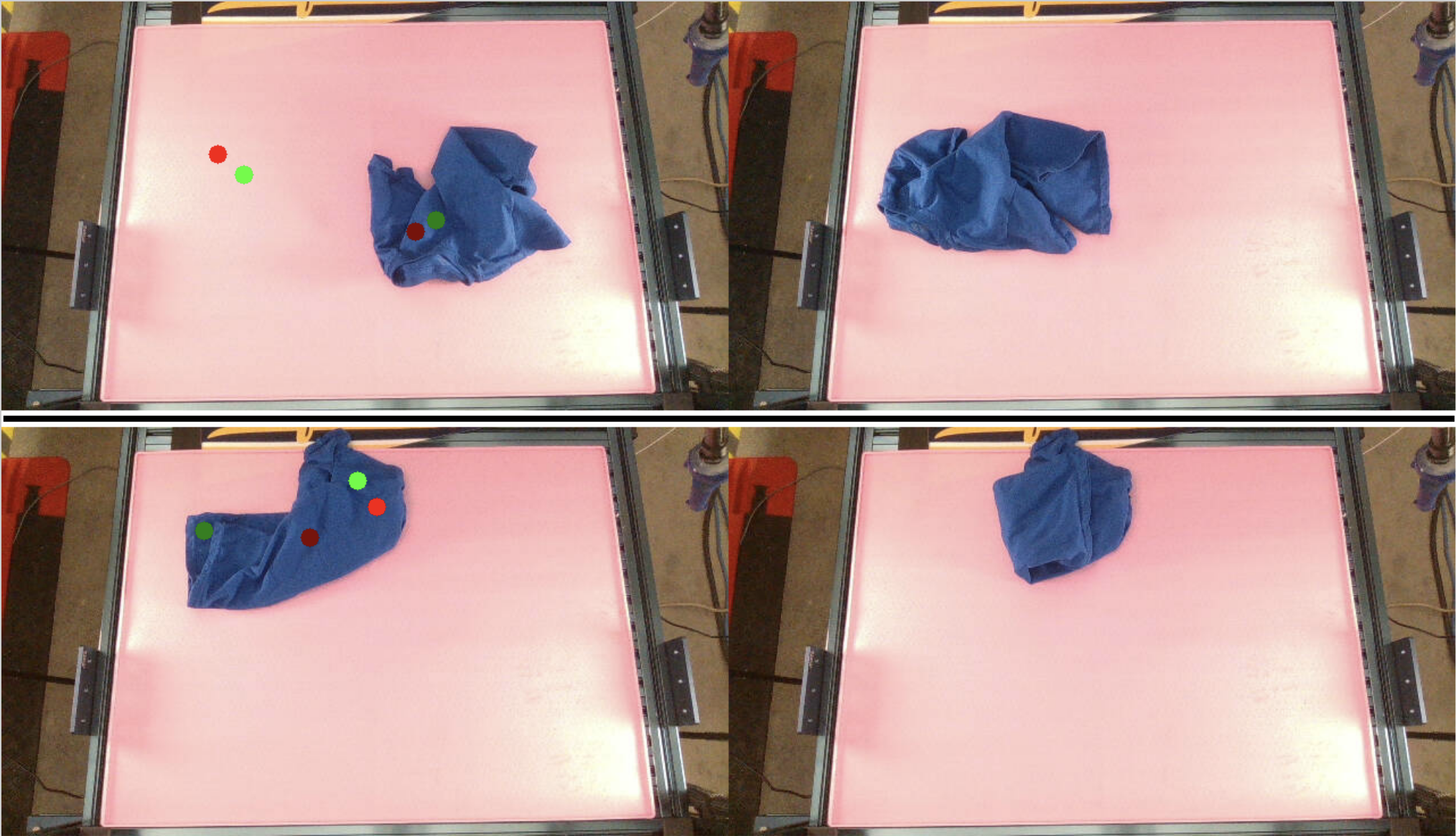}
    \caption{IDYN model predictions on the test set. Each of the two rows is an individual transition where the left image is taken at time $t$ and the right image is taken at time $t + 1$. Ground truth pick-and-place points are in green (dark green for pick, light green for place) and predictions are in red (dark for pick, light for place). Transitions involving translations of the fabric are modeled well (top; mean $\ell_2$ pixel distance 24.9) while more rare transitions such as folding are more challenging (bottom; mean $\ell_2$ pixel distance 60.9).}
    \label{fig:idyn}
\end{figure}

We train the network for 100 epochs with a batch size of 16 and an Adam optimizer with learning rate $1e-4$ and $\ell_2$ regularization $1e-5$, saving the model weights with the lowest test loss. For stability the observation and action values are normalized to $[0,1]$, and the 640 $\times$ 360 images are center cropped to 360 $\times$ 360. After training, the $\ell_2$ distance between predicted points and ground truth points are 51.1 $\pm$ 22.8 pixels (10.0 $\pm$ 4.5\% of maximum error) on the test set; see Figure~\ref{fig:idyn} for examples. At test time, the first input to the network is the current observation and the second input is an observation of the fully flattened shirt. The network output specifies the pick-and-place action to execute on the system. In this algorithm and other algorithms with model outputs, if the pick point misses the fabric, it is analytically corrected to the nearest point on the fabric mask.

\subsubsection{CRL} Here we use the same dataset as the previous section. We train a CNN to predict the scalar change in pixel coverage $\textrm{cover}(\mathbf{o}_{t+1}^m) - \textrm{cover}(\mathbf{o}_t^m)$ from inputs $(\mathbf{o}_t, \mathbf{a}_t)$. Similar to IDYN, we use a ResNet-34 with pretrained weights to encode $\mathbf{o}_t$ into a 1000-dimensional vector. We also tile the action 250$\times$ to match the dimension of the encoded image, which was found to improve performance in this work and prior work \cite{wu2020learning}. After concatenating the two vectors, the state is passed through two fully connected layers of size $(2000,256)$ and $(256,1)$ respectively with ReLU activation and hyperbolic tangent output activation to produce the predicted change in coverage. The CNN has a total of 22.3 million parameters.

As in IDYN, we train for 100 epochs with a batch size of 16 and an Adam optimizer with learning rate $1e-4$ and $\ell_2$ regularization $1e-5$, saving the model weights with the lowest test loss. For stability, observation and action values are normalized to $[0,1]$ and the delta coverage values are normalized to $[-1,1]$. The 640 $\times$ 360 images are center cropped to 360 $\times$ 360. After training, the predicted coverage deltas are 2603 $\pm$ 2230 pixels (2.6 $\pm$ 2.2\% of maximum error) off from the ground truth labels on the test set. At test time, we randomly sample 10,000 pick-and-place actions on the current observation with the RAND strategy and select the action that results in the highest network output. The forward passes are batched into 100 observation-action pairs at a time to decrease total inference time. We also implemented action sampling with the Cross Entropy Method (CEM) but found that this did not significantly change the output actions.

\subsubsection{LP$_0$AP$_1$}\label{ssec:appLPAP} 

To train the imitation learning model, we first collect a dataset of 108 human-labeled pick points on the T-shirt. The pick points are collected through the GUI in Section~\ref{ssec:gui}, which shows the analytically computed place points to the human teleoperator before the actions are executed. The dataset is augmented with online data via 3 iterations of DAgger \cite{dagger}, with potentially multiple pick points labeled per individual observation.

We train an Fully Convolutional Network (FCN) \cite{FCNs} with a ResNet-34 backbone to output a heatmap corresponding to pixel-wise pick affordances, using a sigmoid activation function at the final layer and binary cross-entropy loss. We center-crop the input images to 320 $\times$ 320. To provide a smooth target for the neural network, we add Gaussian distributions ($\sigma$ = 8 pixels) around each labeled pick point. While training, we augment the dataset using the following sequential operations in a random order:
\begin{enumerate}
    \item Flip left/right with probability 50\%.
    \item Flip up/down with probability 50\%.
    \item Rotate $r_k 90^{\circ}$ where $r_k \sim \mathcal{U}(\{0, 1, 2, 3\})$.
    \item Apply a continuous rotation of $r_k \sim \mathcal{U}(-90, 90)$.
    \item Apply an affine transform with x and y scales in the range $(0.9, 1.1)$, translation percentages of $5\%$.
.\end{enumerate}

We train for 100 epochs with a batch size of 4, using an Adam optimizer with a learning rate of $1e-4$. The total number of trainable parameters in the network is 22.8 million.

At runtime, we sample a pick point from the thresholded (value $> 0.1$) heatmap with sampling probability proportional to the intensity of the output pixel value. The place point is computed the same way as is done in AEP.

\subsubsection{KP}
We collect a training dataset with human-annotated keypoints for the collar, sleeve midpoints, and base corners of the shirt (when discernible). We first collect and label 305 images with the RAND policy to train an initial KP policy; we execute this policy to collect an additional 376 images and labels that better represent the distribution of encountered states. We train the same FCN as the previous subsection with the same hyperparameters but have \textit{three} output heatmaps (instead of one) for the collar, sleeves, and base points. These heatmaps are transformed into a series of points by thresholding the normalized heatmaps with a lower bound of $0.2$, with each class of points restricted to an upper-bound of 1, 2, and 2 points for the collar, sleeve, and base points, respectively.

To achieve the optimal transform between the current and template flattened T-shirt, we compute $\argmin\limits_\theta \|p_{obs} - R_\theta p_{template}\|_2$ %constrained by each point observation matching with its corresponding class (reference sleeve points matched with observed sleeve points, etc.). 
where $p_{obs}$ specifies the $SO(2)$ pose of the T-shirt, $R_\theta$ is the 2D rotation matrix corresponding to a rotation of $\theta$, and $p_{template}$ specifies the $SO(2)$ pose of the template T-shirt centered at the visual center of mass of the observed t-shirt. The $\|\cdot\|_2$ cost is computed over the \textit{visible keypoints} and their corresponding points on the template T-shirt. Using the best-match template and observation keypoints, we find the keypoint pair with the largest $\ell_2$ error and execute a pick-and-place action to move the T-shirt keypoint to its target location on the template.

\subsubsection{DROP}

The DROP algorithm is a variant of the LP$_0$AP$_1$ algorithm. Here, we attempt to intelligently combine a dynamic primitive ($\textrm{drop}$) with the algorithm to reduce the number of steps required, as follows:

\begin{enumerate}
    \item Define $\mathcal{S}_d$ as a discretized state space of coverage values that evenly divides the possible coverage values into 200 bins. Define a hierarchical action space $\mathcal{A}_d = \{\textrm{LP}_0\textrm{AP}_1, \textrm{drop}\}$.
    \item Model the transition dynamics of each action in $\mathcal{A}_d$. For LP$_0$AP$_1$, model $$P(s'|s, \textrm{LP}_0\textrm{AP}_1) = s + P(\Delta s | \textrm{LP}_0\textrm{AP}_1)$$
    Boundary cases are handled via clipping and re-normalizing the distribution. 
    For $\textrm{drop}$ actions, we model 
    $$P(s'|s, \textrm{drop}) = P(s'| \textrm{drop})$$
    Intuitively, we model the coverage after a drop as independent of the previous state and the change in coverage from pick-and-place actions as conditional on the previous state. We estimate these quantities in a data-driven manner by executing rollouts of $\textrm{LP}_0\textrm{AP}_1$ or $\textrm{drop}$ and recording the $(s, s')$ tuples.
    \item Once we obtain a full matrix of transition probabilities $P(s'|s, a)$, with $s, s' \in \mathcal{S}_d, a \in \mathcal{A}_d$, we run tabular Q-iteration using Bellman backups, no discount factor, and the reward function below to encourage reaching a high-coverage state as quickly as possible.
    $$r(s, a) = 
        \begin{cases}
          -1 & s < C \\
          0 & s \geq C \\
       \end{cases}
    $$
    Transition dynamics at $s \geq C$ are modified to remain stationary.
    \item Once we obtain an optimal policy from Q-iteration, we run this on the robot. In practice, the policy ends up in the form
    $$\pi(s) = 
        \begin{cases}
        \textrm{drop} & s < T \\
        \textrm{LP}_0\textrm{AP}_1 & s \geq T \\
       \end{cases}
    $$
    Intuitively, the stochastic drop can be interpreted as a geometric series with some probability of escaping the low-coverage regime, after which it is optimal to execute pick-and-place actions so as to prevent eliminating progress in high coverage states. Q-iteration returns $T = 45\%$.
\end{enumerate}

\subsection{Folding Algorithm Details}\label{ssec:algdetails2}

\subsubsection{HUMAN}\label{ssec:appfHUMAN} As described in the main text, this baseline allows the human to freely specify both pick and place points via the graphical user interface in Section~\ref{ssec:gui}.

\subsubsection{ASM} Analytic Shape-Matching requires only a single human demonstration. The human is shown a template image $\mathcal{T}$ of the flattened T-shirt and selects pick and place points for a folding sequence. These points are specified only once for a given folding sequence.

During execution, we compute the best match of the template with the observation $\mathbf{o}_{\textrm{t}}^m$ as follows: %, which we expect to be a rotated version of the t-shirt in the workspace. Because the camera is overhead, we can approximate this well as a pixel-space transformation of each point on the shirt $p$ as $$p' = R_\theta p + T $$

$$\argmax_\theta %\in \{0^{\circ}, 1^{\circ}, ..., 359^{\circ}\}} 
 \ \texttt{sum}(\texttt{and} (\mathbf{o}_t^m, R_{\theta} \mathcal{T} + (\textrm{com}(\mathbf{o}_t^m) - \textrm{com}(\mathcal{T})))) $$ 

where $R_\theta$ is the 2D rotation matrix corresponding to a rotation of $\theta$, $\texttt{and}(\cdot)$ is the pixelwise AND operation, and $(\textrm{com}(\mathbf{o}_t^m) - \textrm{com}(\mathcal{T}))$ is a translation from the template shirt's center of mass to that of the observation. We then transform each demonstration point $p$ with the resulting transform (i.e., $R_\theta p + (\textrm{com}(\mathbf{o}_t^m) - \textrm{com}(\mathcal{T}))$) to get the corresponding action.
%Finally, for each user-selected point $p$ relative to the template, we calculate $Rp + T$ to calculate the pick and place points with respect to the current shirt. 
Before executing actions we project pick points and place points onto the shirt mask to avoid missed grasps.

\subsubsection{LP$_0$LP$_1$} \label{ssec:appfHUMAN}
\begin{figure}
    \centering
    \includegraphics[width=0.45\textwidth]{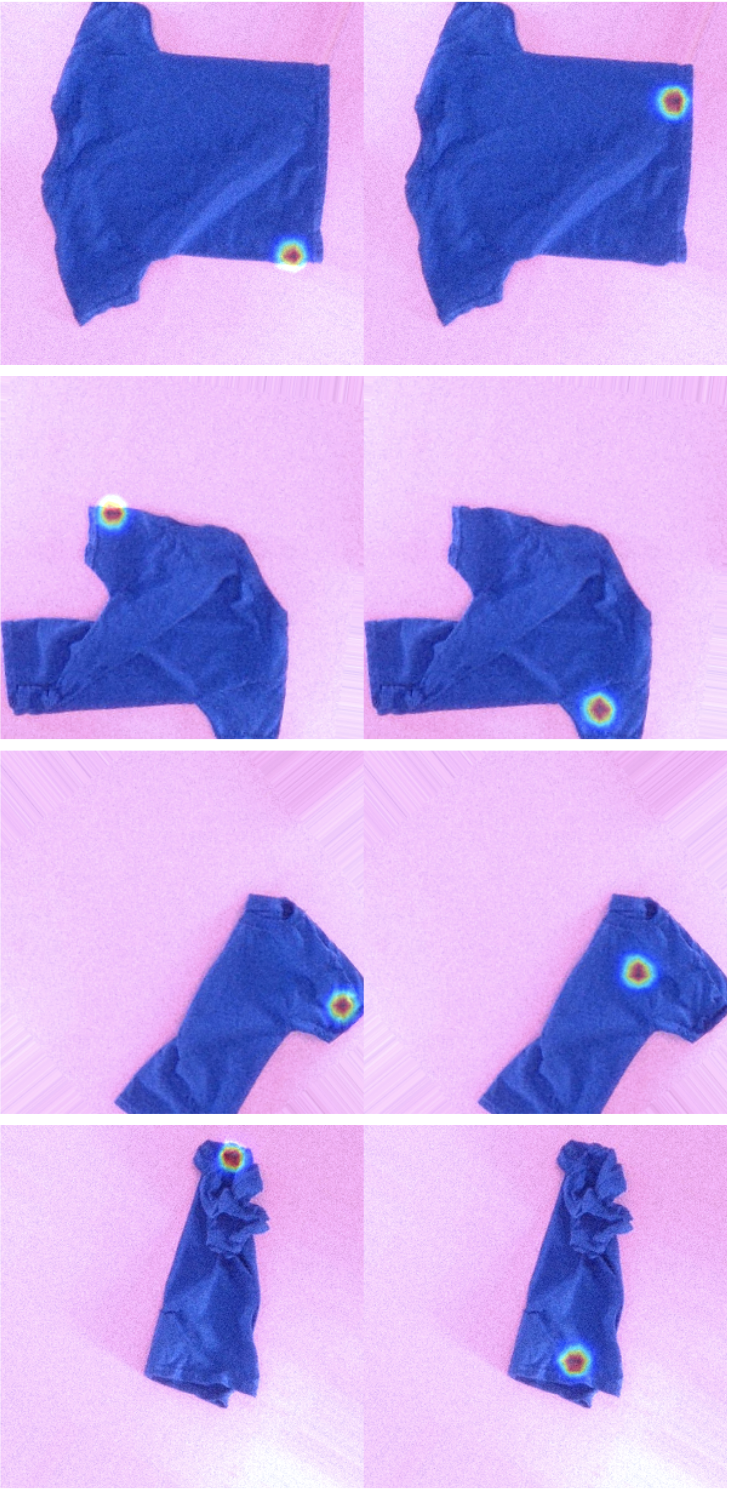}
    \caption{LP$_0$LP$_1$ keypoint predictions on the test set (i.e., unseen affine transforms). Each row is a different data sample where the left image shows the predicted pick and the right shows the predicted place. The 4 steps together comprise a folding demonstration.}
    \label{fig:LPLP}
\end{figure}

We collect 2 folding demonstrations with 4 pick-and-place actions each via human teleoperation. We augment the dataset by a factor of 20 to get a total of 160 data points to reduce overfitting and build robustness to rotations and translations. Specifically, we perform the following affine transforms in a random order:
\begin{enumerate}
    \item Rotate a random multiple of 90 degrees (i.e., 0, 90, 180, or 270)
    \item Scale the image in $x$ uniformly at random between  95\% and 105\%
    \item Scale the image in $y$ uniformly at random between 95\% and 105\%
    \item Rotate uniformly at random between -45 degrees and 45 degrees
    \item Translate in $x$ uniformly at random between -5\% and 5\%
    \item Translate in $y$ uniformly at random between -5\% and 5\%
\end{enumerate}
We also crop the images around the visual center of mass $\textrm{com}(\cdot)$ (without crossing the workspace bounds) to encourage further translational invariance. We then train and run inference with the same FCN as LP$_0$AP$_1$ but predict 2 output keypoints instead of 1 (i.e., pick and place point). See Figure~\ref{fig:LPLP} for test set predictions. At test time we execute the model outputs and terminate after 4 actions, though the termination condition may be learned in future work to allow more closed-loop behavior.

\subsubsection{A-ASM} This baseline simply executes ASM after the flattening is performed by LP$_0$AP$_1$ instead of HUMAN.

\subsection{Graphical User Interface}\label{ssec:gui}

\begin{figure}
    \centering
    \includegraphics[width=0.49\textwidth]{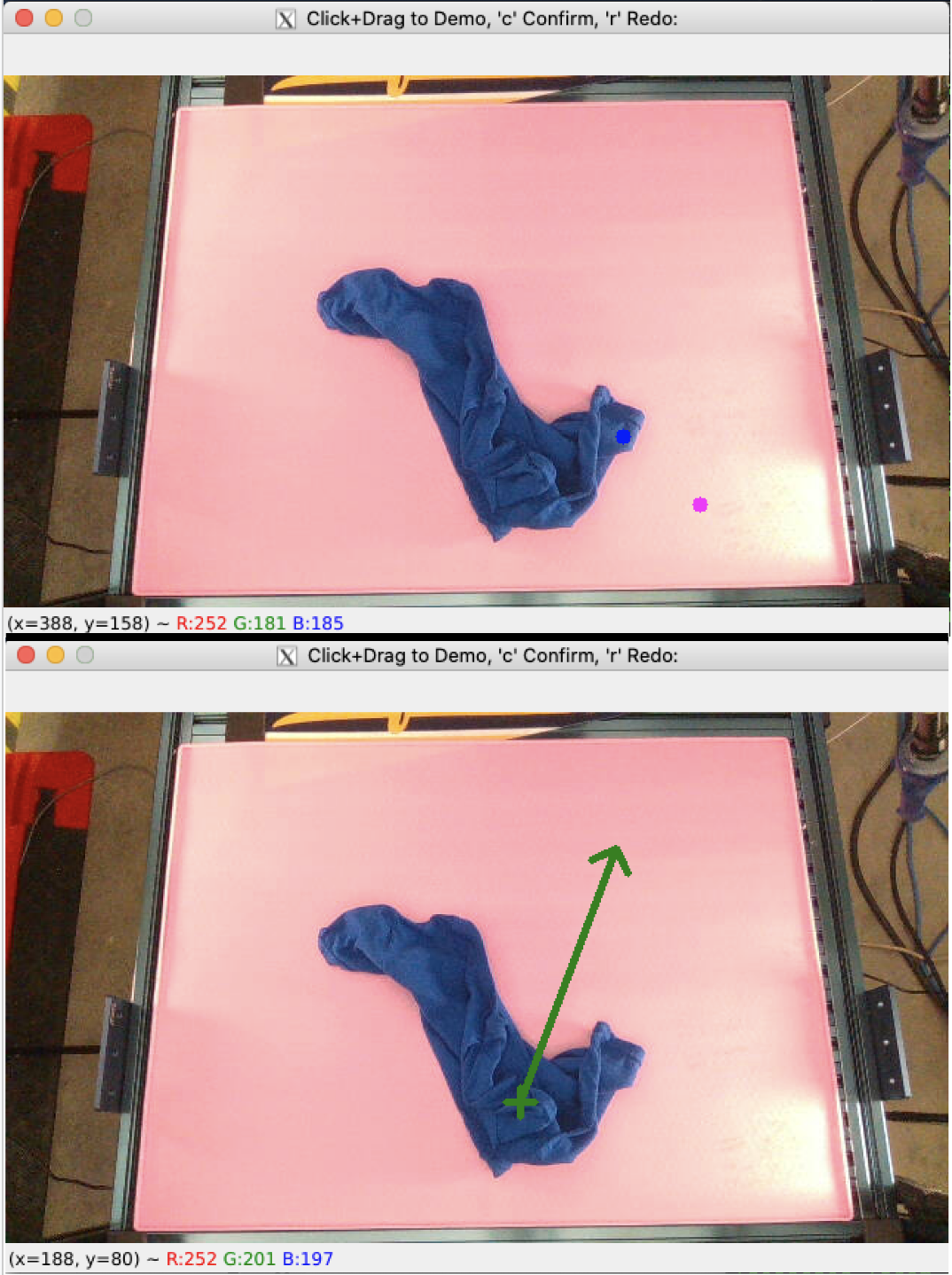}
    \caption{\textbf{Top:} The user clicks to select a pick point and the place point is automatically computed with the strategy in Section~\ref{ssec:appAEP}. \textbf{Bottom:} The user clicks to select a pick point, drags the mouse, and releases to select a place point.}
    \label{fig:gui}
\end{figure}

To enable human teleoperation, we develop a simple graphical user interface (GUI) with OpenCV (\url{https://opencv.org/}). The GUI displays the overhead RGB camera image of the current garment state and allows the user to specify pixels for the pick and/or place point with the mouse (Figure~\ref{fig:gui}). These points are deprojected into 3D world coordinates through the known depth camera transform and parameterize a pick-and-place action for the robot to execute. The place points are analytically computed for LP$_0$AP$_1$ and human-specified for LP$_0$LP$_1$, HUMAN flattening, and HUMAN folding. The GUI is used to provide demonstration data for the former two algorithms and execute actions for the latter two. Human labels of visible keypoints are collected with a similar interface for the KP algorithm.

\subsection{Action Primitive Details}\label{ssec:primitives}

\subsubsection{Flattening Pick-and-Place} A flattening pick-and-place primitive is parameterized by a 2D pixel pick point $p_0$ and a 2D pixel place point $p_1$. It consists of the following steps, where all steps but the first are calls to the \texttt{step()} function of the PyReach Gym API:
\begin{itemize}
    \item Deproject the pick point $p_0$ into the world coordinates ($x_1, y_1, z_1$) of the top layer of the fabric at $p_0$ via the known depth camera transform. Deproject the place point $p_1$ into world coordinates ($x_2, y_2, z_2$).
    \item Move the gripper to a fixed initial pose in the center of the workspace about 0.2 meters above the worksurface with the gripper oriented top-down and the jaw open.
    \item Translate the gripper to $(x_1, y_1)$ without changing the height or rotation.
    \item Lower the gripper to $z_1$.
    \item Close the gripper to grasp the fabric.
    \item Raise the gripper to 0.1 meters above its current height.
    \item Translate the gripper to $(x_2, y_2)$ without changing the height.
    \item Open the gripper to release the fabric.
    \item Move the arm out of the field of view before the next image observation is captured.
\end{itemize}
All arm movements are commanded with joint velocity limits 1.04 radians per second and joint acceleration limits 1.2 radians per seconds squared.

\subsubsection{Folding Pick-and-Place} A folding pick-and-place primitive is also parameterized by a 2D pixel pick point $p_0$ and a 2D pixel place point $p_1$. It consists of the following steps, where differences from the flattening motion are in bold:
\begin{itemize}
    \item Deproject the pick point $p_0$ into the world coordinates ($x_1, y_1, z_1$) of the top layer of the fabric at $p_0$ via the known depth camera transform. Deproject the place point $p_1$ into world coordinates ($x_2, y_2, z_2$).
    \item Move the gripper to a fixed initial pose in the center of the workspace about 0.2 meters above the worksurface with the gripper oriented top-down and the jaw open.
    \item Translate the gripper to $(x_1, y_1)$ without changing the height or rotation.
    \item Lower the gripper to \textbf{at most 0.1 meters below $z_1$, but not below the height of the worksurface.}
    \item Close the gripper to grasp the fabric.
    \item Raise the gripper to 0.1 meters above its current height.
    \item Translate the gripper to $(x_2, y_2)$ without changing the height \textbf{at a lower speed than the other motions.}
    \item \textbf{Lower the gripper to $z_1$.}
    \item Open the gripper to release the fabric.
    \item \textbf{Raise the gripper to 0.1 meters above $z_1$.}
    \item Move the arm out of the field of view before the next image observation is captured.
\end{itemize}
All arm movements but the translation to $(x_2, y_2)$ are commanded with joint velocity limits 1.04 radians per second and joint acceleration limits 1.2 radians per seconds squared. The translation has a joint velocity limit slower than the other motions by a factor of four (i.e., 0.26 radians per second).

\subsubsection{COM Drop} 
A center-of-mass (COM) drop motion is parameterized by pick point $p_0$, which is automatically computed as the visual COM of the fabric. Differences from the flattening pick-and-place motion are in bold.
\begin{itemize}
    \item Deproject the pick point $p_0$ into the world coordinates ($x_1, y_1, z_1$) of the top layer of the fabric at $p_0$ via the known depth camera transform.
    \item Move the gripper to a fixed initial pose in the center of the workspace about 0.2 meters above the worksurface with the gripper oriented top-down and the jaw open.
    \item Translate the gripper to $(x_1, y_1)$ without changing the height or rotation.
    \item Lower the gripper to $z_1$.
    \item Close the gripper to grasp the fabric.
    \item \textbf{Move the gripper to 0.3 meters above the center of the workspace.}
    \item Open the gripper to release the fabric.
    \item Move the arm out of the field of view before the next image observation is captured.
\end{itemize}

\subsubsection{Reset} A `crumple' or reset operation for flattening performs the following motion 6 consecutive times:
\begin{itemize}
    \item Move the arm to a fixed initial pose in the center of the workspace about 0.2 meters above the worksurface with the gripper oriented top-down and the jaw open.
    \item Compute a random $x$ and $y$ offset, each uniformly sampled between -0.1 and 0.1 (For reference, the workspace is about 0.7m by 0.5m.).
    \item Move the gripper to the center of the workspace at the height of the workspace, but offset in $x$ and $y$ as above.
    \item Close the gripper.
    \item Move the gripper to 0.3 meters above the center of the workspace.
    \item Open the gripper.
\end{itemize}

\subsubsection{Recenter}\label{ssec:prim-recenter}

During flattening rollouts, if the visual center of mass is sufficiently far away from the center of the workspace in $x$ or $y$ (by about 100 pixels in each direction, where the full image is 640 $\times$ 360), we perform a recentering primitive. This is a flattening pick-and-place primitive where $p_0$ is the pixel nearest to the center of the workspace with a small amount of noise applied ($\pm$ 12.5 pixels in $x$ and $y$) and $p_1$ is the center of the workspace. 

\subsubsection{Recovery}\label{ssec:prim-recovery}

We perform a random recovery action if no flattening progress is being made, i.e., coverage is below 75\%, at least 5 actions have been executed, and the last 2 actions achieve lower mean coverage than the preceding 2 actions. Specifically, we select a random $p_0$ on the shirt mask and compute $p_1$ with AEP.

%\subsubsection{Flinging}

\subsection{Ablation Studies}\label{ssec:ablations}

We run ablation studies on \lpap{} to evaluate the efficacy of various components. We find that multiple design choices, when eliminated, cause the average number of actions required to flatten the shirt to increase dramatically.

\subsubsection{Recovery Actions}
For this experiment, we disable the random recovery actions we take when we detect a lack of progress in coverage (Section \ref{ssec:prim-recovery}).
\subsubsection{Recentering Actions}
Here, we remove the recentering actions we take when the shirt's visual center of mass is too far from the center of the workspace (Section \ref{ssec:prim-recenter}).
\subsubsection{Action Shrinking}
Here, we disable the reduction in action magnitudes when the shirt reaches higher coverage states (Section \ref{ssec:appAEP}).

\begin{figure}[!ht]
    \centering
    \includegraphics[width=0.49\textwidth]{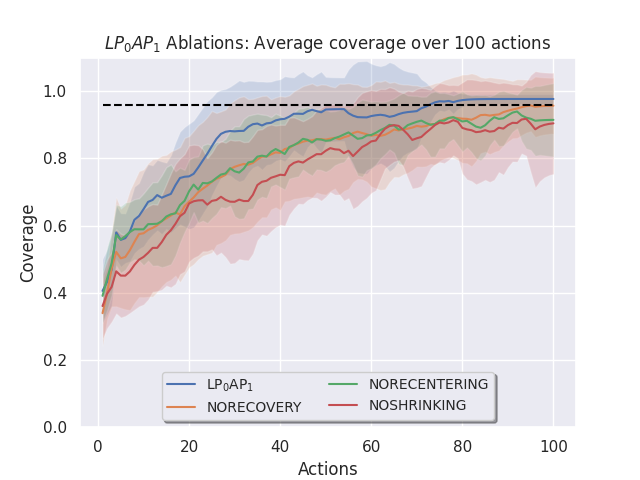}
    \caption{Coverage vs. time for each of the ablation experiments. Shading represents one standard deviation, and the horizontal dashed line is the flattening succcess threshold (96\%). All ablations converge less quickly than LP$_0$AP$_1$.}
    \label{fig:ablation_coverages}
\end{figure}

The ablation experiments demonstrate that the additional primitives we introduce indeed improve the performance of \lpap{}. Figure \ref{fig:ablation_coverages} and Table \ref{tab:ablations} illustrate the slower convergence of \lpap{} without each of these primitives and optimizations to a fully flattened t-shirt state. These experiments show that for practical behavior cloning involving deformable objects, adding such manipulation primitives can be beneficial in accelerating progress to the goal state.

\begin{table}[t]
\caption{
\small
Flattening ablation results. We report maximum coverage, number of actions, number of samples in the dataset, and evaluation time, where averages and standard deviations are computed over 10 trials.
}
\centering
\begin{tabular}{l | l l r r}\label{tab:ablations}
\textbf{Algorithm} & \textbf{\% Coverage} & \textbf{Actions} & \textbf{Dataset} & \textbf{Time/Act (s)} \\ \hline 
% \todo{fill out the table with the real numbers}\\
LP$_0$AP$_1$ & \textbf{97.7 $\pm$ 1.4} & \textbf{31.9 $\pm$ 17.2} & 524 & 25.6 $\pm$ 0.9  \\
No Recovery & 96.0 $\pm$ 7.3 & 53.5 $\pm$ 30.3 & 524 & 25.5 $\pm$ 1.2 \\
No Recentering & 96.4 $\pm$ 1.0 & 65.6 $\pm$ 27.2 & 524 & 25.6 $\pm$ 1.2 \\
No Shrinking & 94.8 $\pm$ 6.6 & 61.2 $\pm$ 28.9 & 524 & 25.5 $\pm$ 0.8 \\
\end{tabular}
%\vspace*{-10pt}
\label{tab:analytic}
\end{table}

\end{document}